\documentclass[10pt,journal,compsoc]{IEEEtran}

\usepackage{array}
\usepackage{epsfig}
\usepackage{graphicx}
\usepackage{amsmath}
\usepackage{amssymb}
\usepackage{xcolor}
\usepackage{multirow, makecell}
\usepackage{lineno}
\usepackage{enumerate}
\usepackage{amsfonts}
\usepackage{gensymb}
\usepackage{bm}
\usepackage{diagbox}
\usepackage{tabularx}
\usepackage{rotating}
\newcolumntype{L}[1]{>{\raggedright\let\newline\\\arraybackslash\hspace{0pt}}m{#1}}
\newcolumntype{C}[1]{>{\centering\arraybackslash}m{#1}}
\newcolumntype{R}[1]{>{\raggedleft\let\newline\\\arraybackslash\hspace{0pt}}m{#1}}
\usepackage{xmpmulti}
\usepackage{arydshln}
\usepackage[nocompress]{cite}
\usepackage{url}

\usepackage{subfigure}

\begin{document}

\title{Learning to Discriminate Information for Online Action Detection: Analysis and Application}

\author{Sumin~Lee,
Hyunjun~Eun,
Jinyoung~Moon,
Seokeon~Choi, 
Yoonhyung~Kim,
Chanho~Jung,
and~Changick~Kim,~\IEEEmembership{Senior~Member,~IEEE}

\IEEEcompsocitemizethanks{\IEEEcompsocthanksitem {S. Lee, S. Choi, and C. Kim are with the School of Electrical Engineering, Korea Advanced Institute of Science and Technology (KAIST), Daejeon 34141, Republic of Korea. (e-mail: \{suminlee94, seokeon, changick\}@kaist.ac.kr)}
\IEEEcompsocthanksitem {H. Eun is with the AI Service Division, SK Telecom, Seoul, Republic of Korea. (e-mail: hyunjun.eun@sk.com)}
\IEEEcompsocthanksitem {J. Moon and Y. Kim are with the Electronics and Telecommunications Research Institute (ETRI), Daejeon 34129, Republic of Korea. Also, J.Moon is also with the ICT department, the University of Science and Technology (UST). (e-mail: \{jymoon, yhkim1127\}@etri.re.kr)}
\IEEEcompsocthanksitem {C. Jung is with the Department of Electrical Engineering, Hanbat National University, Daejeon 34158, Republic of Korea. (e-mail: peterjung@hanbat.ac.kr)}
\IEEEcompsocthanksitem {Corresponding author: Chanho Jung}
}
\thanks{}}

\markboth{}%
{}

\IEEEtitleabstractindextext{
\begin{abstract}
Online action detection, which aims to identify an ongoing action from a streaming video, is an important subject in real-world applications.
For this task, previous methods use recurrent neural networks for modeling temporal relations in an input sequence.
However, these methods overlook the fact that the input image sequence includes not only the action of interest but background and irrelevant actions.
This would induce recurrent units to accumulate unnecessary information for encoding features on the action of interest.
To overcome this problem, we propose a novel recurrent unit, named Information Discrimination Unit (IDU), which explicitly discriminates the information relevancy between an ongoing action and others to decide whether to accumulate the input information.
This enables learning more discriminative representations for identifying an ongoing action.
In this paper, we further present a new recurrent unit, called Information Integration Unit (IIU), for action anticipation.
Our IIU exploits the outputs from IDN as pseudo action labels as well as RGB frames to learn enriched features of observed actions effectively.
In experiments on TVSeries and THUMOS-14, the proposed methods outperform state-of-the-art methods by a significant margin in online action detection and action anticipation. 
Moreover, we demonstrate the effectiveness of the proposed units by conducting comprehensive ablation studies.
\end{abstract}

}

\maketitle

\IEEEdisplaynontitleabstractindextext
\IEEEpeerreviewmaketitle

\IEEEraisesectionheading{
\section{Introduction}
\label{sec:introduction}}
\IEEEPARstart{T}{emporal} action detection~\cite{chao2018cvpr,liu2019cvpr,xu2017iccv,zhang2019aaai,zhao2017iccv} has been widely studied in an offline setting, which allows making a decision for the detection after fully observing a long, untrimmed video.
This is called offline action detection.
In contrast, online action detection aims to identify ongoing actions from streaming videos at every moment in time.
This task is useful for many real-world applications, such as autonomous driving \cite{kim2019cvpr}, robot assistants \cite{koppula2013iros}, and surveillance systems \cite{iwashita2013bmvc, shu2015cvpr}.
Also, online action detection in these applications can be developed to more challenging tasks (e.g., action anticipation).
\begin{figure}[t!]
\centering{\includegraphics[width=.99\linewidth]{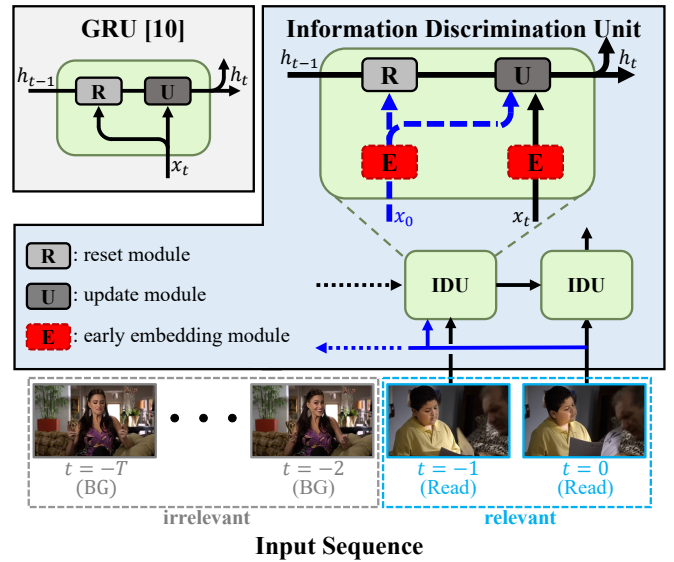}}
\caption{\label{fig1}Comparison between GRU \cite{cho2014emnlp} and the proposed Information Discrimination Unit (IDU) for online action detection.
Our IDU extends GRU with two novel components, a mechanism utilizing current information (blue lines) and an early embedding module (red dash boxes).
First, reset and update modules in our IDU additionally take the current information (i.e., $x_0$) to consider whether the past information (i.e, $h_{t-1}$ and $x_t$) are relevant to an ongoing action such as $x_0$.
Second, the early embedding module is introduced to consider the relation between high-level features for both information.
}
\end{figure}

For online action detection, recent methods~\cite{gao22017bmvc, xu2019iccv} employ recurrent neural networks (RNNs) with recurrent units (e.g., long short-term memory (LSTM)~\cite{hochreiter1997nc} and gated recurrent unit (GRU)~\cite{cho2014emnlp}) to effectively model temporal sequences. 
To exploit discriminative representations, they introduce additional modules.
Gao~\textit{et. al}~\cite{gao22017bmvc} propose a reinforcement module to output correct predictions as early as possible with sequence-level rewards, and Xu~\textit{et. al}~\cite{xu2019iccv} presented a recurrent module that considers the temporal correlations between current and future actions.
However, these methods overlook the fact that the given input video contains not only the ongoing action but also background and irrelevant actions.
Specifically, the conventional recurrent units accumulate the input information without explicitly considering its relevance to the current action, and thus the learned representation would be less discriminative.
Note that, in the task of detecting actions online, ignoring such a characteristic of streaming videos makes the problem more challenging \cite{geest2016eccv}.

In this paper, we investigate on the question of \textit{how RNNs can learn to explicitly discriminate relevant information from irrelevant information for detecting actions in the present}.
To this end, we propose a novel recurrent unit that extends GRU~\cite{cho2014emnlp} with the following two components: 1) a mechanism utilizing current information and 2) an early embedding module (see Fig. \ref{fig1}).
We name our recurrent unit Information Discrimination Unit (IDU).
Specifically, reset and update modules in our IDU learn the relationship between an ongoing action and past information (i.e., $x_t$ and $h_{t-1}$) by additionally taking current information (i.e., $x_0$) at every time step.
We further introduce the early embedding module to model the relation more effectively .
By adopting action classes and feature distances as supervisions, our early embedding module learns the high-level features of the current and past actions.
Based on IDU, our Information Discrimination Network (IDN) effectively determines whether to use input information in terms of its relevance to the current action.
This enables the network to learn a more discriminative representation for detecting ongoing actions.

According to recent online action detection studies~\cite{xu2019iccv, gao22017bmvc}, online action detection has a close relationship with action anticipation.
Those two tasks take a streaming video as an input, but predict actions at different points in time.
In this paper, to apply the outputs of IDN as psuedo action labels for action anticipation, we further introduce a new recurrent unit, called Information Integration Unit (IIU).
Our IIU takes not only RGB frames but also action labels as input to exploit action-relevant features on observed actions.
Based on our IIU, Information Integration Network (IIN) captures enriched and contextual information for predicting unseen future actions.

We perform extensive experiments on two benchmark datasets (i.e., TVSeries~\cite{geest2016eccv} and THUMOS-14~\cite{jiang2015url}).
Our IDN for online action detection achieves state-of-the-art performances of 86.1\% mcAP and 60.3\% mAP on TVSeries~\cite{geest2016eccv} and THUMOS-14~\cite{jiang2015url}, respectively.
These performances significantly outperform those performances of TRN \cite{xu2019iccv}, the previous best performer, by 2.4\% mcAP on TVSeries and 13.1\% mAP on THUMOS-14.
We also evaluate the action anticipation performance of our IIN forecasting future action after $t$ seconds.
Our IIN performs better than state-of-the-art methods~\cite{xu2019iccv,gao22017bmvc} by achieving 74.3\% mcAP on TVSeries and 37.1\% mAP on THUMOS-14 for predicting actions after 2 seconds.
These results on action anticipation show that the proposed relation modeling with IDU and IIU is effective, and that our IDN have a great potential to be broadly applied beyond a single task (i.e., online action detection).
Moreover, we conduct comprehensive ablation studies of two proposed units.
Throughout these ablation studies, we demonstrate that taking additional information suitable for each task with the sophisticatedly designed recurrent unit structure is effective to obtain desired information from videos.

The main contributions of this paper are, as follows:
\renewcommand\labelitemi{\tiny$\bullet$}
\begin{itemize}
\item Different from previous methods, we investigate on how recurrent units can explicitly discriminate relevant information from irrelevant information for online action detection.
\item We introduce a novel recurrent unit, IDU, with a mechanism using current information at every time step and an early embedding module to effectively model the relevance of input information to an ongoing action.
\item We further present a new recurrent unit, IIU, for the task of action anticipation. By employing the outputs of IDN as pseudo action labels, our IIU generates contextual features by integrating two different modal inputs (i.e., visual and pseudo action label features) of each time step.
\item We evaluate the performance of online action detection and action anticipation on two benchmark datasets. Experimental results show that IDN and IIN outperform the state-of-the-art methods by a large margin. Moreover, we demonstrate the effectiveness of the proposed method by conducting comprehensive ablation studies.
\end{itemize}

This paper is an extended version of our previous conference paper~\cite{eun2020learning}.

\section{Related Work}
\label{sec:rw}
\subsection{Offline Action Detection}
\label{sec:rw-offline}
The goal of offline action detection is to detect the start and end times of action instances from fully observed long untrimmed videos.
Most methods \cite{chao2018cvpr,shou2016cvpr,zhao2017iccv, dai2017iccvr,xu2017iccv, shou2017cvpr, r2-3, r2-2,r2-6} consist of two steps including action proposal generation~\cite{r2-1, r2-4} and action classification~\cite{r2-7, r2-5}.
Shou \emph{et al.}~\cite{shou2016cvpr} introduced a multi-stage Segment-CNN framework that consists of proposal, classification, and localization networks.
The proposal network eliminates uncertain candidate segments, and the localization network adjusts each action instance to have higher temporal overlaps with the ground truths. 
Dai \emph{et al.}~\cite{dai2017iccvr} proposed TCN for determining the ranking of proposals.
To measure the start and end times of proposals, TCN explicitly considers the local context information of each proposal.
SSN \cite{zhao2017iccv} first evaluates actionness scores for temporal locations to generate temporal intervals.
Then, these intervals are classified by modeling the temporal structures and completeness of action instances.
Xu \emph{et al.}~\cite{xu2017iccv} introduced R-C3D, which consists of three-dimensional fully convolutional networks.
By sharing convolution features of the proposal and the classification pipeline, the computation efficiency of R-C3D is improved, and end-to-end training is possible.
TAL-Net \cite{chao2018cvpr}, including the proposal generation and classification networks, is the extended version of Faster R-CNN \cite{ren2015nips} for offline action detection.
Inspired by TAL-Net, PCG-TAL~\cite{r2-3} is proposed, which take advantages of complimentarity between the anchor-based and frame-based paradigms.
Lin \emph{et al.}~\cite{r2-2} presented first purely anchor-free temporal localization method with boundary pooling for generating fine-grained predictions.
CDC~\cite{shou2017cvpr} predicts frame-level dense prediction and precise boundaries of action segments by simultaneously performing temporal up-sampling and spatial down-sampling.
To avoid frame-level annotations that are unsuitable for real-world scenarios, some studies~\cite{3r_iccv, 3r_aaai, r2-6} have explored weakly supervised temporal action localization, in which video-level annotations.
FAC-Net~\cite{3r_iccv} explores bilateral relations between action and foreground with the foreground-action consistency in order to discriminate foreground and background.
Huang~\emph{et al.}~\cite{3r_aaai} proposed a prototypical network for action-background and action-action separations.
Su~\emph{et al.}~\cite{r2-6} proposed a two-stage approach to generate high-quality frame-level pseudo labels.

Other methods \cite{donahue2015cvpr,yeung2018ijcv} with LSTM have been also studied for per-frame prediction.
Donahue \emph{et al.}~\cite{donahue2015cvpr} show that LSTM-style RNNs can produce significant improvements in visual time-series modeling.
Yeung \emph{et al.}~\cite{yeung2018ijcv} proposed a variant of LSTM, called MultiLSTM, for modeling temporal relations between multiple and dense labels.

\subsection{Early Action Prediction}
\label{sec:rw-early}
This task is similar to online action detection but focuses on recognizing actions from the partially observed videos.
Many methods~\cite{hoai2012cvpr, hoai2014ijcv,ma2016cvpr, early1, early2, early3, early4} have been developed to detect an ongoing action as early as possible.
Hoai and la Torre \cite{hoai2012cvpr, hoai2014ijcv} introduced the problem of early action prediction for the first time. 
They designed a maximum-margin framework with the extended structured SVM \cite{tsochantaridis2005jmlr} to accommodate sequential data.
Ma \emph{et al.}~\cite{ma2016cvpr} proposed the modified training loss based on ranking losses on the detection score and detection score margin.
The first loss on the detection score constrains the detection score to be monotonically non-decreasing.
The second loss on the detection score margin between a correct action class and all others forces the margin to be monotonically non-decreasing.
Cai \emph{et al.} \cite{cai2019aaai} proposed to transfer action knowledge learned from full videos to partially observed videos for the prediction of partial videos.

\subsection{Online Action Detection}
\label{sec:rw-online}
Given a streaming video, online action detection aims to identify actions as soon as each video frame arrives, without observing future video frames.
Geest \emph{et al.} \cite{geest2016eccv} defined the problem of online action detection in detail, and introduced a new large dataset, TVSeries.
They also analyzed the performance changes with a variation in viewpoint, occlusion, truncation, and compared several baseline methods on TVSeries dataset.
For a fair comparison, an evaluation protocol for online action detection is defined in~\cite{geest2016eccv}.
In their later work~\cite{geest2018wacv}, a two-stream feedback network with LSTMs is introduced to individually perform the interpretation of the features and the modeling of the temporal dependencies.
Gao, Yang, and Nevatia~\cite{gao22017bmvc} proposed Reinforced Encoder-Decoder (RED) network with a reinforcement loss.
The encoder-decoder network of RED uses an LSTM network, and the reinforcement module is additionally proposed to consider sequence-level rewards.
The reward function of RED encourages the network to make correct decisions as early as possible.
They designed RED for the task of action anticipation, which aims to predict future actions after a few seconds.
However, RED can perform online action detection by setting the anticipation time to 0.
Xu \emph{et al.}~\cite{xu2019iccv} introduced Temporal Recurrent Network (TRN) that predicts future information and utilizes the predicted future as well as the past and current information together for detecting a current action.
A TRN cell, which consists of a temporal decoder, a future gate, and a spatio-temporal accumulator, exploits the temporal correlations between current and future actions.
Xu~\emph{et al.}~\cite{3r_nips} presented a transformer-based network, named Long Short-term Transformer (LSTR), to jointly model long- and short-term temporal relationships.

Aforementioned methods \cite{geest2016eccv,gao22017bmvc,xu2019iccv} for online action detection adopt RNNs to model a current action sequence.
However, the RNN units, such as LSTM \cite{hochreiter1997nc} and GRU~\cite{cho2014emnlp} operate without explicitly considering whether input information is relevant to the ongoing action or not.
Therefore, the current action sequence is modeled based on both relevant and irrelevant information of current actions, which results in a less discriminative representation.

\subsection{Action Anticipation}
\label{sec:aa}
The goal of action anticipation is to forecast the class and the duration (i.e., when each action will start and end) of future actions.
Early works have investigated anticipations of the immediate next action after the observation~\cite{next1, next2, next4, next3, next5, next6}.
Lan \emph{et al.}~\cite{next1} introduced a new representation called the hierarchical movemes, which are captured from the typical structure of human behavior.
Mahmud \emph{et al.}~\cite{next2} proposed a network based on a hybrid Siamese network to jointly train the future label and the starting time.
Koppula \emph{et al.}~\cite{next3} represented an anticipatory temporal conditional random field (ATCRF) to models rich spatial-temporal relations with object affordances.
In~\cite{next4}, DARKO is proposed for forecasting future behaviors by incorporating an online inverse reinforcement learning approach.
Pei \emph{et al.}~\cite{next5} proposed an unsupervised learning algorithm and an event parsing algorithm for inferring the goal of the agent and predicting their plausible intended actions.
Wang~\emph{et al.}~\cite{3r_tip} introduced a recurrent encoder-decoder network that predicts future human motion with the use of pose velocities and temporal positional embeddings.

Later, methods for predicting longer time horizon of a few seconds are investigated~\cite{fewsec1, gao22017bmvc, xu2019iccv, fewsec5}.
Vondrick \emph{et al.}~\cite{fewsec1} introduced a framework for anticipating human actions and objects in unlabeled videos.
They used unlabeled videos to learn to estimate the visual representation in the future and then apply recognition algorithms on the predicted future features. 
Qi \emph{et al.}~\cite{fewsec5} designed a method to predict future actions from partially observed RGB-D videos. 
Gao \emph{et al.}~\cite{gao22017bmvc} utilized a reinforcement learning scheme with a LSTM encoder-decoder architecture. 
In~\cite{xu2019iccv}, TRN, which is designed to predict the near future for online action detection, is leveraged for forecasting the next action after 2 seconds.

These methods only consider RGB frame sequences as input and usually consist of two steps: 1) understanding observed actions and 2) forecasting future actions.
By using both frame and corresponding action label sequences, networks can relieve the burden of the first step, enabling them to focus the second step.
In the end, to exploit both RGB frame and action label sequences for action anticipation, we introduce a recurrent unit that integrates action-relevant information on two different modality sequences to enriched features of observed actions.
Note that, for assuming real-world environments, we utilize the outputs of our IDN for action anticipation instead of ground truths.
This also demonstrates the applicability of our IDU and IDN.

Unlike the aforementioned methods, some of works~\cite{ego2, ego5, fewsec4, ego3, ego4,ego1,ego6} focused on predicting actions from ego-centric videos.
While generic videos contain the full body movements of an actor, ego-centric videos capture an actor's hand and objects that the actor is interacting with.
For ego-centric action anticipation, Dessalene \emph{et al.}~\cite{ego2} proposed an anticipation module, which generates hand-object contact map and next-active object segmentation. 
In~\cite{ego5}, an object manipulation graph is proposed to model relations between hands and objects.
Furnari \emph{et al.}~\cite{fewsec4} proposed an architecture with two LSTMs to summarize the past and formulated ego-centric action predictions. 
In~\cite{ego3}, a Slow-Fast LSTM model is proposed to extract slow and fast feature from three different modalities (i.e., RGB, optical flow and extracted objects).
Fernando \emph{et al.}~\cite{ego4} proposed Jaccard vector similarity to correlate past features with the future.



\begin{figure*}[t!]
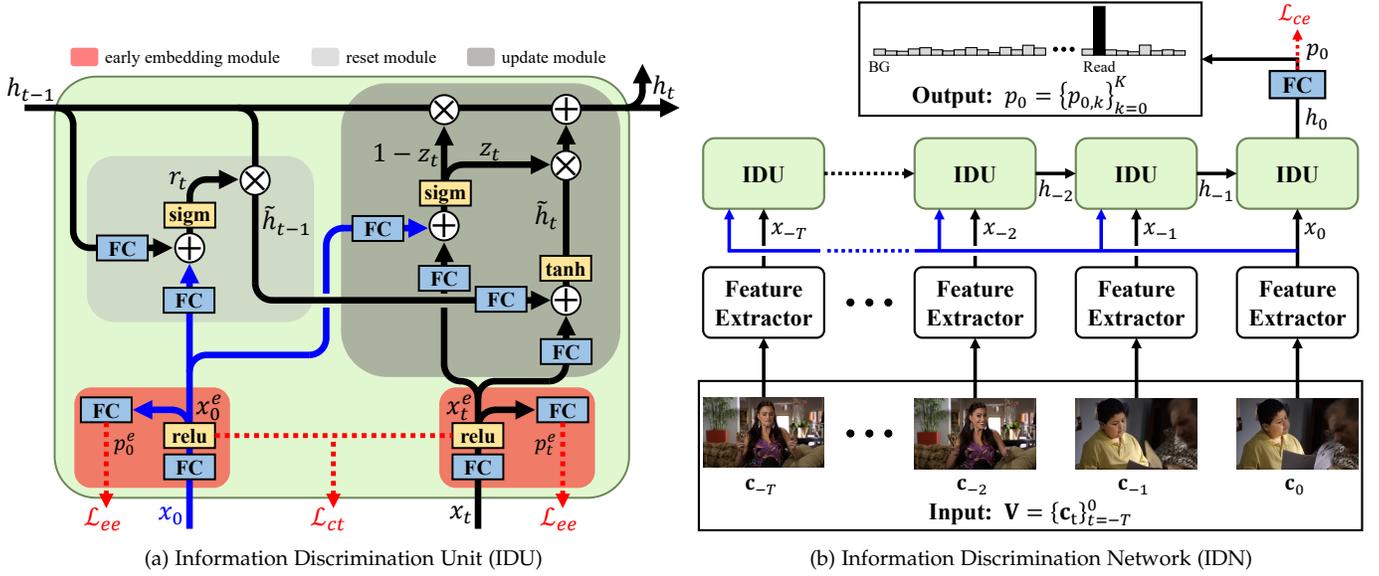

\centering
\begin{minipage}[t]{0.499\linewidth}
{\centering{\includegraphics[width=.99\linewidth]{./figures/fig2-b.pdf}}}
\centering{\footnotesize{(a) Information Discrimination Unit (IDU)}}
\end{minipage}
\begin{minipage}[t]{0.495\linewidth}
{\centering{\includegraphics[width=.99\linewidth]{./figures/fig2-a.pdf}}}
\centering{\footnotesize{(b) Information Discrimination Network (IDN)}}
\end{minipage}
\vspace{0.05cm}
\caption{\label{fig2}Illustration of our Information Discrimination Unit (IDU) and Information Discrimination Network (IDN).
(a) Our IDU extends GRU with two new components, a mechanism using current information (i.e., $x_0$) (blue lines) and an early embedding module (red boxes). The first encourages reset and update modules to model the relation between past information (i.e., $h_{t-1}$ and $x_t$) and an ongoing action. The second enables IDU to effectively model the relation between high-level features for the input information.
(b) Given an input streaming video $\mathbf{V}=\{\mathbf{c}_t\}_{t=-T}^0$ consisting of sequential chunks, IDN models a current action sequence and outputs the probability distribution $p_0$ of the current action over $K$ action classes and background.
}
\end{figure*}
\section{Preliminary: Gated Recurrent Unit}
\label{sec:gru}
We first analyze GRU \cite{cho2014emnlp} to compare differences between the proposed IDU and GRU.
GRU is one of the recurrent units, which is much simpler than LSTM.
Two main components of GRU are reset and update gates.

The reset gate $r_t$ is computed based on a previous hidden state $h_{t-1}$ and an input $x_t$, as follows:
\begin{eqnarray}
\label{eq:gru1}
r_t = \sigma (\textbf{W}_{hr}h_{t-1} + \textbf{W}_{xr}x_t),
\end{eqnarray}
where $\textbf{W}_{hr}$ and $\textbf{W}_{xr}$ are parameters to be trained and $\sigma$ is the logistic sigmoid function.
Then, the reset gate determines whether a previous hidden state $h_{t-1}$ is ignored as
\begin{eqnarray}
\label{eq:gru2}
\tilde{h}_{t-1} = r_t \otimes h_{t-1},
\end{eqnarray}
where $\tilde{h}_{t-1}$ is a new hidden state at time $t-1$, and $\otimes$ indicates the element-wise multiplication.

Similar to $r_t$, the update gate $z_t$ is also computed based on $h_{t-1}$ and $x_t$ as
\begin{eqnarray}
\label{eq:gru3}
z_t = \sigma (\textbf{W}_{xz}x_t + \textbf{W}_{hz}h_{t-1}),
\end{eqnarray}
where $\textbf{W}_{xz}$ and $\textbf{W}_{hz}$ are learnable parameters.
The update gate decides whether a hidden state $h_t$ is updated with a new hidden state $\tilde{h}_t$, as follows:
\begin{eqnarray}
h_t = (1-z_t) \otimes h_{t-1} + z_t \otimes \tilde{h}_t,
\end{eqnarray}
where
\begin{eqnarray}
\tilde{h}_t = \eta (\textbf{W}_{x\tilde{h}}x_t + \textbf{W}_{\tilde{h}\tilde{h}}\tilde{h}_{t-1}).
\end{eqnarray}
Here $\textbf{W}_{x\tilde{h}}$ and $\textbf{W}_{\tilde{h}\tilde{h}}$ are trainable parameters and $\eta$ is the tangent hyperbolic function.

Based on reset and update gates, GRU effectively drops and accumulates information to learn a compact representation.
However, there are limitations when we applied GRU to online action detection as below:

First, the past information, including $x_t$ and $h_{t-1}$, directly affects the decision of the reset and update gates.
For online action detection, the relevant information to be accumulated is the information related to the current action.
Thus, it is advantageous to make a decision based on the relation between the past information and the current action instead.
To this end, we reformulate the computations of the reset and update gates by additionally taking the current information (i.e., $x_0$) as input. 

This enables the reset and update gates to drop the irrelevant information and accumulate the relevant information regarding the ongoing action.
Second, it is implicitly considered that the input features of the reset and update gates represent valuable information.
We augment GRU with an early embedding module with supervisions, action classes, and feature distances, so that the input features explicitly describe actions.
By optimizing features for the target task and dataset, our early embedding module also lets the reset and update gates focus on accumulating the relevant information along with the recurrent steps.
We discuss the effect of the early embedding module in Sec~\ref{sec:ee}.

\section{Approach}
\label{sec:approach}
We first describe our IDU in detail and then explain on IDN for online action detection.
In Fig.~\ref{fig2}, the schematic view of our IDU and the framework of IDN are illustrated.
Next, we introduce our IIU and IIN for action anticipation, which are described in Fig.~\ref{fig3}.

\subsection{Information Discrimination Unit}
Our IDU extends GRU with two new components: a mechanism utilizing current information (i.e., $x_0$) and an early embedding module.
We explain IDU with early embedding, reset, and update modules, which takes a previous hidden state $h_{t-1}$, the features at each time $x_t$, and the features at current time $x_0$ as input and outputs a hidden state $h_t$ (see Fig. \ref{fig2} (a)).
\subsubsection{Early Embedding Module.}
Our early embedding module individually processes the features at each time $x_t$ and the features at current time $x_0$ and outputs embedded features $x_t^e$ and $x_0^e$, as follows:
\begin{eqnarray}
x_t^e = \zeta(\textbf{W}_{xe}x_t), \\
x_0^e = \zeta(\textbf{W}_{xe}x_0),
\end{eqnarray}
where $\bm{W}_{xe}$ is a weight matrix and $\zeta$ is the ReLU \cite{nari2010icml} activation function.
Note that we share $\bm{W}_{xe}$ for $x_t$ and $x_0$.
We omit a bias term for simplicity.

To encourage $x_t^e$ and $x_0^e$ to represent specific actions, we introduce two supervisions: action classes and feature distances.
First, we process $x_t^e$ and $x_0^e$ to obtain probability distributions $p_t^e$ and $p_0^e$ over $K$ action classes and background:
\begin{eqnarray}
p_t^e = \xi(\textbf{W}_{ep}x_t^e), \\
p_0^e = \xi(\textbf{W}_{ep}x_0^e),
\end{eqnarray}
where $\bm{W}_{ep}$ is a shared weight matrix to be learned and $\xi$ is the softmax function.
We design a classification loss ${\cal{L}}_{ee}$ by adopting the multi-class cross-entropy loss as
\begin{eqnarray}
{\cal{L}}_{ee} = -\sum_{k=0}^K \left( \ y_{t,k} \text{log}(p_{t,k}^e) + y_{0,k} \text{log}(p_{0,k}^e) \right),
\end{eqnarray}
where $y_{t,k}$ and $y_{0,k}$ are ground truth labels.
Second, we use the contrastive loss \cite{chopra2005cvpr,hadsell2006cvpr} proposed to learn an embedding representation by preserving the distance between similar data points close and dissimilar data points far on the embedding space in metric learning \cite{sohn2016nips}.
By using $x_t^e$ and $x_0^e$ as a pair, we design our contrastive loss ${\cal{L}}_{ct}$ as
\begin{eqnarray}
\label{eq11}
\begin{aligned}
{\cal{L}}_{ct} = & \textbf{1} \{ y_t=y_0 \} D^2(x_t^e, x_0^e) \\
& + \textbf{1} \{ y_t \neq y_0 \} \text{max}(0, m-D^2(x_t^e, x_0^e)),
\end{aligned}
\end{eqnarray}
where $D^2(a,b)$ is the squared Euclidean distance and $m$ is a margin parameter.

We train our embedding module with ${\cal{L}}_{ee}$ and ${\cal{L}}_{ct}$, which provides more representative features for actions.
More details on training will be provided in Section 4.2.

\subsubsection{Reset Module}
Our reset module takes the previous hidden state $h_{t-1}$ and the embedded features $x_0^e$ to compute a reset gate $r_t$ as 
\begin{eqnarray}
\label{eq12}
r_t = \sigma(\textbf{W}_{hr}h_{t-1} + \textbf{W}_{x_0r}x_0^e),
\end{eqnarray}
where $\textbf{W}_{hr}$ and $\textbf{W}_{x_0r}$ are weight matrices which are learned.
We define $\sigma$ as the logistic sigmoid function same as GRU.
We then obtain a new hidden state at time $t-1$ $\tilde{h}_{t-1}$, as follows:
\begin{eqnarray}
\tilde{h}_{t-1} = r_t \otimes h_{t-1}.
\end{eqnarray}

Different from GRU, we compute the reset gate $r_t$ based on $h_{t-1}$ and $x_0^e$.
This enables our reset gate to effectively drop or take the past information according to its relevance to an ongoing action.
\begin{figure*}[t]
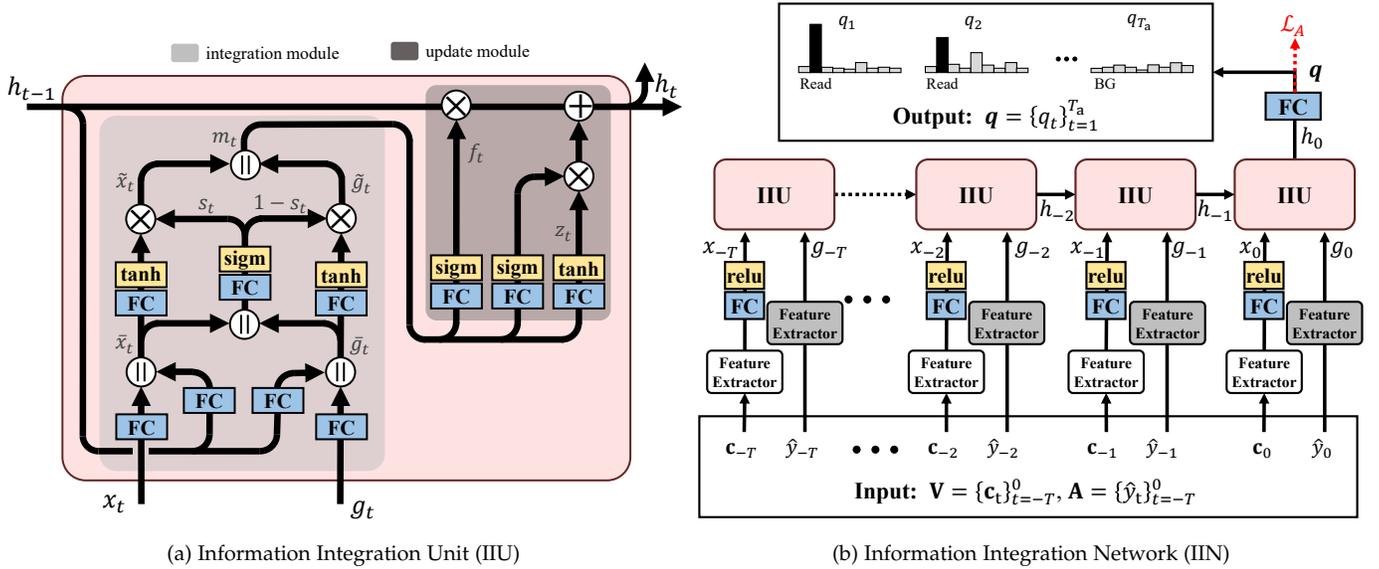

\centering
\begin{minipage}[t]{0.499\linewidth}
{\centering{\includegraphics[width=.99\linewidth]{./figures/fig3-a.pdf}}}
\centering{\footnotesize{(a) Information Integration Unit (IIU)}}
\end{minipage}
\begin{minipage}[t]{0.495\linewidth}
\centering{\includegraphics[width=.99\linewidth]{./figures/fig3-b.pdf}} 
\centering{\footnotesize{(b) Information Integration Network (IIN)}}
\end{minipage}
\vspace{0.05cm}
\caption{\label{fig3}Illustration of our Information Integration Unit (IIU) and Information Integration Network (IIN).
(a) Our IIU integrates two different modality sequences $x_t$ and $g_t$, considering a previous integrated state $h_{t-1}$. IIU consists of an integration module for combining two input sequences and an update module for updating the current integrated state $h_t$.
(b) Given an input video $\mathbf{V}=\{\mathbf{c}_t\}_{t=-T}^0$ and corresponding pseudo action labels $\mathbf{A} = \{\hat{y}_{t}\}_{t=-T}^{0}$, IIN captures historical and contextual information and predicts the probability distributions $\mathbf{Q}=\{q_t\}_{t=1}^{T_{a}}$ of future actions.
}
\end{figure*}
\subsubsection{Update Module}
Our update module adopts the embedded features $x_t^e$ and $x_0^e$ to compute an update gate $z_t$, as follows:
\begin{eqnarray}
\label{eq14}
z_t = \sigma(\textbf{W}_{x_tz}x_t^e + \textbf{W}_{x_0z}x_0^e),
\end{eqnarray}
where $\textbf{W}_{x_tz}$ and $\textbf{W}_{x_0z}$ are trainable parameters.
Then, a hidden state $h_t$ is computed, as follows:
\begin{eqnarray}
h_t = (1-z_t) \otimes h_{t-1} + z_t \otimes \tilde{h}_t,
\end{eqnarray}
where 
\begin{eqnarray}
\label{eq16}
\tilde{h}_t = \eta (\textbf{W}_{x_t\tilde{h}}x_t^e + \textbf{W}_{\tilde{h}\tilde{h}}\tilde{h}_{t-1}).
\end{eqnarray}
Here $\tilde{h}_t$ is a new hidden state and $\eta$ is the tangent hyperbolic function.
$\textbf{W}_{x_t\tilde{h}}$ and $\textbf{W}_{\tilde{h}\tilde{h}}$ are trainable parameters.

There are two differences between the update modules of our IDU and GRU. 
The first difference is that our update gate is computed based on $x_t^e$ and $x_0^e$.
This allows the update gate to consider whether $x_t^e$ is relevant to an ongoing action.
Second, our update gate uses the embedded features which are more representative in terms of specific actions.

\subsection{Information Discrimination Network}
In this section, we explain our recurrent network, called IDN, for online action detection (see Fig. \ref{fig2}.(b)).

\subsubsection{Problem Setting}
To formulate the online action detection problem, we follow the same setting as in previous methods \cite{gao22017bmvc,xu2019iccv}.
Given a streaming video $\textbf{V}=\{ \textbf{c}_t \}_{t=-T}^0$ including current and $T$ past chunks as input, our IDN outputs a probability distribution $p_0= \{ p_{0,k}\}_{k=0}^K$ of a current action over $K$ action classes and background.
Here we define a chunk $c= \{ I_n \}_{n=1}^N$ as the set of $N$ consecutive frames.
$I_n$ indicates the $n$-th frame.

\subsubsection{Feature Extractor} \label{sec:feat_ext}
We use TSN \cite{wang2016eccv} as a feature extractor.
TSN takes an individual chunk $\textbf{c}_t$ as input and outputs an appearance feature vector $x_t^a$ and a motion feature vector $x_t^m$.
We concatenate $x_t^a \in \mathbb{R}^{d_{a}}$ and $x_t^m \in \mathbb{R}^{d_{m}}$ into a two-stream feature vector $x_t=[x_t^a \: \| \: x_t^m]\in \mathbb{R}^{d_{x}}$, where $[\cdot\|\cdot]$ indicates a vector concatenation operation.
Here $d_x$ equals to $d_a + d_m$.
After that, we sequentially feed $x_t$ and $x_0$ into our IDU.

\subsubsection{Training}
We feed the hidden state $h_0$ at current time into a fully connected layer to obtain the final probability distribution $p_0$ of an ongoing action, as follows:
\begin{eqnarray}
p_0 = \xi(\textbf{W}_{hp}h_0),
\end{eqnarray}
where $\textbf{W}_{hp}$ is a trainable matrix and $\xi$ is the softmax function.

We define a classification loss ${\cal{L}}_{ce}$ for a current action by employing the standard cross-entropy loss as
\begin{eqnarray}
{\cal{L}}_{ce} = -\sum_{t=-T}^0 \sum_{k=0}^K \ y_{t,k} \text{log}(p_{t,k}),
\end{eqnarray}
where $y_{t,k}$ are the ground truth labels for the $t$th time step.
We train our IDN by jointly optimizing ${\cal{L}}_{ce}$, ${\cal{L}}_{ee}$, and ${\cal{L}}_{ct}$ by designing a multi-task loss ${\cal{L}}_{OAD}$, as follows:
\begin{eqnarray}
\label{eq18}
{\cal{L}}_{OAD} = {\cal{L}}_{ce} + \alpha ( {\cal{L}}_{ee} + {\cal{L}}_{ct} ),
\end{eqnarray}
where $\alpha$ is a balance parameter.

\subsection{Information Integration Unit and Network}
\subsubsection{Information Integration Unit}
We explain our new recurrent unit, IIU, in this section. 
Our IIU exploits pseudo action label sequences obtained from IDN as well as visual sequences. 
This strategy enables our IIU to learn enriched features from the two sequences with different properties of information.
Concretely, the use of the action labels assists to extract action-relevant features from visual information effectively.
%
For these two input feature sequences, the proposed IIU integrates visual and action label features with previous hidden state features. 
This encourages to effectively exploit comprehensive understanding of historical action information for forecasting unseen future actions.

As described in Fig.~\ref{fig3} (a), our IIU consists of two modules: 1) an integration module for assimilating a visual feature vector $x_t$ and an action label feature vector $g_t$ to an integrated feature vector $m_t$ and 2) an update module for updating a previous hidden state $h_{t-1}$ to a hidden state $h_t$ based on $m_t$.
We obtain $g_t$ from pseudo action label $\hat{y}_{t}$.
The details about how to generate $x_t$ and $g_t$ are described in Sec.~\ref{sec:IIN}.

In the integration module, the updated visual features $\bar{x_t}$ and the label features $\bar{g_t}$ are computed by using $h_{t-1}$, as follows:
\begin{align}
\bar{x}_{t} = [ &\: \mathbf{W}_{x\bar{x}} {x}_{t} \:  \| \:  \mathbf{W}_{h\bar{x}} h_{t-1} \: ],\\
\bar{g}_{t} = [ &\: \mathbf{W}_{g\bar{g}} g_{t} \:  \|  \: \mathbf{W}_{h\bar{g}} h_{t-1} \: ],
\end{align}
where $\mathbf{W}_{x\bar{x}}$, $\mathbf{W}_{h\bar{x}}$, $\mathbf{W}_{g\bar{g}}$, and $\mathbf{W}_{h\bar{g}}$ are trainable weight matrices.
To determine how much each modality representation is activated, we then compute fusion scores $s_t$ as
\begin{align}
s_{t} = \sigma(\mathbf{W}_{s} & [ \:  \bar{x}_{t} \:  \| \:  \bar{g}_{t} \: ]),
\end{align}
where $\mathbf{W}_{s}$ is a trainable weight matrix.
Next, the weighted visual features $\tilde{x_t}$ and the action label features $\tilde{g_t}$ are calculated such that:
\begin{align}
&\tilde{x}_{t} = s_{t} \: \otimes \:  \eta (\mathbf{W}_{\bar{x}\tilde{x}} \bar{x}_{t}),\\
\tilde{g}_{t} &= (1-s_{t}) \: \otimes \:  \eta (\mathbf{W}_{\bar{g}\tilde{g}} \bar{g}_{t}),
\end{align}
where $\mathbf{W}_{\bar{x}\tilde{x}}$ and $\mathbf{W}_{\bar{g}\tilde{g}}$ are learnable weight matrices.
By concatenating these two feature vectors, we obtain the integrated features $m_{t}$, as follows:
\begin{align}
    m_{t} = [\:\tilde{x}_{t}\: || \: \tilde{g}_{t} \:]\:.
\end{align}

In the update module, the previous hidden state $h_{t-1}$ is updated based on $m_{t}$ to the current hidden state $h_t$.
We define a forget gate $f_t$ and and an update gate $z_t$ to determine how much $h_{t-1}$ is ignored and how much $m_t$ contributes to $h_t$, respectively.
The hidden state $h_t$ is computed as:
\begin{align}
h_{t} = f_{t} \otimes h_{t-1} + z_{t} \otimes \eta(\mathbf{W}_{mh}m_{t}),
\end{align}
where
\begin{align}
    f_{t} = \sigma(\mathbf{W}_{mf} m_{t}), \\
    z_{t} = \sigma(\mathbf{W}_{mz} m_{t}).
\end{align}
Here $\mathbf{W}_{mh}$, $\mathbf{W}_{mf}$ and $\mathbf{W}_{mz}$ are trainable parameters.

\subsubsection{Information Integration Network} \label{sec:IIN}
As illustrated in Fig.~\ref{fig3} (b), we propose a new recurrent network, named Information Integration Network (IIN), for action anticipation.

To obtain action label input, we convert the probability distribution $p_t$ of IDN to a one-hot vector $\hat{y}_t$, which has 1 at the maximum $p_t$ value.
Then, we obtain label features $g_t$, as follows:
\begin{align}
g_{t} = G( \, \hat{y}_{t} \, ),
\end{align}
where the function G is the action label feature extractor composed of several linear layers.
For visual feature $x_t$, we employ a linear layer to the output of the feature extractor for dimension reduction, as follows:
\begin{align}
x_{t} = \xi( \mathbf{W}_{x} F(\mathbf{c}_{t})),
\end{align}
where the function $F$ is the feature extractor, and $\mathbf{W}_{x}$ is a learnable parameter.
The feature extractor of IIN is the same one of IDN.

For action anticipation, we feed the hidden state $h_0$ to three fully connected layers to predict the probabilities for $T_a$ future actions $\mathbf{Q}=\{q_t\}_{t=1}^{T_{a}}\in\mathbb{R}^{T_{a}(K+1)}$, as follows:
\begin{align}
    \mathbf{Q} = \xi (\mathbf{W}_{hq}^{3} \zeta(\mathbf{W}_{hq}^{2} \zeta(\mathbf{W}_{hq}^{1} h_{0}))),
\label{eq:q}
\end{align}
where $\mathbf{W}^{1}_{hq}$, $\mathbf{W}^{2}_{hq}$, and $\mathbf{W}^{3}_{hq}$ are learnable parameters, and $q_t = \{q_{t,k}\}^{K}_{k=0}$ is a probability distribution of a future action over $K$ action classes and background.
We define a classification loss ${\cal{L}}_{AA}$ for training IIN, which is the standard cross-entropy loss, as follows:
\begin{eqnarray}
{\cal{L}}_{AA} = -\sum_{t=1}^{T_a} \sum_{k=0}^{K} \ y_{t,k} \text{log}(q_{t,k}).
\end{eqnarray}

\section{Experimental Settings}
\subsection{Datasets}
\subsubsection{TVSeries}
This dataset~\cite{geest2016eccv} includes 27 untrimmed videos on six popular TV series, divided into 13, 7, and 7 videos for training, validation, and test, respectively.
Each video contains a single episode, approximately 20 minutes or 40 minutes long.
The dataset is temporally annotated with 30 realistic actions (e.g., open door, read, eat, \textit{etc}).
The TVSeries dataset is challenging due to diverse undefined actions, multiple actors, heavy occlusions, and a large proportion of non-action frames.

\subsubsection{THUMOS-14}
The THUMOS-14 dataset~\cite{jiang2015url} consists of 200 and 213 untrimmed videos for validation and test sets, respectively.
This dataset has temporal annotations with 20 sports actions (e.g., diving, shot put, billiards, \textit{etc}).
Each video includes 15.8 action instances and 71$\%$ background on average.
As done in \cite{gao22017bmvc,xu2019iccv}, we used the validation set for training and the test set for evaluation.

\subsection{Evaluation Metric}
For evaluating performance in online action detection, existing methods \cite{geest2016eccv,gao22017bmvc,xu2019iccv} measure mean average precision (mAP) and mean calibrated average precision (mcAP) \cite{geest2016eccv} in a frame level.
Both metrics are computed in two steps: 1) calculating the average precision over all frames for each action class and 2) averaging the average precision values over all action classes.

We evaluate the performance in action anticipation with mAP and mcAP.
Following the evaluation protocol of~\cite{xu2019iccv, gao22017bmvc}, we compute mAP and mcAP about predicted future actions after $t$ seconds, where $t\in \left [ 0.25:0.25:2.0 \right ]$.

\subsubsection{mean Average Precision (mAP).}
On each action class, all frames are first sorted in descending order of their probabilities.
The average precision of the $k$th class over all frames is then calculated based on the precision at cut-off $i$ (i.e., on the $i$ sorted frames).
The final mAP is defined as the mean of the AP values over all action classes.

\subsubsection{mean calibrated Average Precision (mcAP).}
It is difficult to compare two different classes in terms of the AP values when the ratios of positive frames versus negative frames for these classes are different.
To address this problem, Geest \emph{et al.} \cite{geest2016eccv} proposed the calibrated precision as
\begin{eqnarray}
\text{cPrec}(i) = \frac{w\text{TP}(i)}{w\text{TP}(i)+\text{FP}(i)},
\end{eqnarray}
where $w$ is a ratio between negative frames and positive frames.
Similar to the AP, the calibrated average precision of the $k$th class over all frames is computed as
\begin{eqnarray}
\text{cAP}_k=\frac{\sum_i \text{cPrec}(i)\textbf{1}(i)}{N_P}.
\end{eqnarray}
Then, the mcAP is obtained by averaging the cAP values over all action classes.

\begin{table}[t!]
\caption{\label{tab1}Specifications of our IDN. $d_x$ is the dimension of the two-stream feature vector $x_t$, and $K+1$ is the number of action and background classes.}
\vspace{-0.3cm}
\centering
\begin{tabular}{C{2.6cm}|C{.8cm}|C{1.2cm}|C{1.9cm}}\hline 
Module & Type & Weight & Size  \\ \hline\hline
\multirow{2}{*}{\shortstack{Early Embedding\\Module}} & FC & $\textbf{W}_{xe}$ & $d_x\times512$  \\
 & FC & $\textbf{W}_{ep}$ & $512\times (K+1)$ \\ \hline
\multirow{2}{*}{\shortstack{Reset\\Module}} & FC & $\textbf{W}_{hr}$ & $512\times512$ \\
 & FC & $\textbf{W}_{x_0r}$ & $512\times512$ \\ \hline
\multirow{4}{*}{\shortstack{Update\\Module}} & FC & $\textbf{W}_{x_tz}$ & $512\times512$ \\
 & FC & $\textbf{W}_{x_0z}$ & $512\times512$ \\
 & FC & $\textbf{W}_{x_t\tilde{h}}$ & $512\times 512$ \\
 & FC & $\textbf{W}_{\tilde{h}\tilde{h}}$ & $512\times 512$ \\ \hline
Classification & FC & $\textbf{W}_{hp}$ & $512\times (K+1)$  \\ \hline
\end{tabular}
\centering
\end{table}
\begin{table}[t]
\caption{\label{tab2}Specifications of our IIN. $d_x$ is the dimension of the two-stream feature vector $x_t$, $d_g$ is the dimension of the action feature $g_t$, $T_a$ is the number of anticipation steps, and $K$ is the number of action classes.}
\vspace{-0.3cm}
\centering
{\begin{tabular}{C{1.8cm}|C{.5cm}|C{2.1cm}|C{2.3cm}}
\hline
Module & Type & Weight & Size \\ \hline \hline
 \shortstack{Dimension\\Reduction} & FC & $W_{x}$ & $d_x \times 2048 $ \\ \hline
\multirow{4}{*}{\shortstack{Integration\\Module}} & FC & $W_{x \bar{x}}$, $W_{g \bar{g}}$ & $2048 \times 2048$ \\
 & FC & $W_{h \bar{x}}$. $W_{h \bar{g}}$ & $2048 \times 2048$ \\
 & FC & $W_{s}$ & $4072 \times 2048$ \\
 & FC & $W_{\bar{x} \tilde{x}}$, $W_{\bar{g} \tilde{g}}$ & $2048 \times 2048$ \\ \hline
\multirow{3}{*}{\shortstack{Update\\Module}} & FC & $W_{mh}$ & $4096 \times 2048$ \\
 & FC & $W_{mf}$ & $4096 \times 2048$ \\ 
  & FC & $W_{mz}$ & $4096 \times 2048$ \\\hline
\multicolumn{1}{l|}{\multirow{3}{*}{Classification}} & FC & $W_{hq}^{1}$ & $2048 \times 1024$ \\
\multicolumn{1}{l|}{} & FC & $W_{hq}^{2}$ & $1024 \times 2048$ \\
\multicolumn{1}{l|}{} & FC & $W_{hq}^{3}$ & $2048 \times (T_{a} (K+1))$ \\ \hline
\end{tabular}}
\centering
\end{table}
\subsection{Implementation Details}
\subsubsection{Problem Setting.}
We use the same setting as the one used in state-of-the-art methods \cite{gao22017bmvc,xu2019iccv}.
On both TVSeries \cite{geest2016eccv} and THUMOS-14 \cite{jiang2015url} datasets, we extract video frames at $24$ fps and set the number of frames in each chunk $N$ to $6$.
We use $16$ chunks (i.e., $T=15$), which are $4$ seconds long, for the input of IDN and IIN.

\subsubsection{Feature Extractor}
We use a two-stream network as a feature extractor of visual RGB frames for both online action detection and action anticipation.
In the two-stream network, one stream encodes appearance information by taking the center frame of a chunk as input, while another stream encodes motion information by processing an optical flow stack computed from an input chunk.
Among several two-stream networks, we employ the TSN model \cite{wang2016eccv} pretrained on the ActivityNet-v1.3 dataset \cite{heilbron2015cvpr}.
Note that this TSN is the same feature extractor as used in state-of-the-art methods \cite{gao22017bmvc, xu2019iccv}.
The TSN model consists of ResNet-200 \cite{he2016cvpr} for an appearance network and BN-Inception \cite{ioffe2015arxiv} for a motion network.
We use the outputs of the {\fontfamily{qcr}\selectfont Flatten\_673} layer in ResNet-200 and the {\fontfamily{qcr}\selectfont global\_pool} layer in BN-Inception as the appearance features $x_t^a$ and motion features $x_t^m$, respectively.
The dimensions of $x_t^a$ and $x_t^m$ are $d_a=2048$ and $d_m=1024$, respectively, and $d_x$ equals to $3072$.

For action anticipation, we use a feature extractor of an action label stream to encode one-hot vectors of pseudo action labels into high-level features with a dimension of $d_g$.
The feature extractor consists of two fully connected layers followed by a batch normalization layer and a non-linear function (e.g., ReLU), respectively.
In the experiments, we set $d_g$ to $128$.
The feature extractor of action labels is trained with our IIN.

\subsubsection{IDN Architecture and Training}
Table~\ref{tab1} provides the specifications of IDN considered in our experiments.
In the early embedding module, we set the number of the hidden units for $\textbf{W}_{xe}$ to $512$.
In the reset module, both weights $\textbf{W}_{hr}$ and $\textbf{W}_{x_0r}$ have 512 hidden units.
In the update module, we use 512 hidden units for $\textbf{W}_{x_tz}$, $\textbf{W}_{x_0z}$, $\textbf{W}_{x_t\tilde{h}}$, and $\textbf{W}_{\tilde{h}\tilde{h}}$.
According to the number of action classes, we set $K+1$ to 31 for TVSeries and 21 for THUMOS-14.

To train our IDN, we use the stochastic gradient descent optimizer with the learning rate of 0.01 for both THUMOS-14 and TVSeries datasets.
We set the batch size to 128 and balance the numbers of action and background samples in terms of the class of $\textbf{c}_0$.
We empirically set the margin parameter $m$ in Eq.~(\ref{eq11}) to $1.0$ and the balance parameter $\alpha$ in Eq.~(\ref{eq18}) to $0.3$.

\subsubsection{IIN Architecture and Training}
In Table~\ref{tab2}, we describe the specification of IIN used in our experiments.
The dimension of all hidden units in IIU is set to $2048$.
Our IIN predicts future actions up to $T_a$ time steps, which is set to 8. 
Like IDN, we set $K+1$ to 31 for TVSeries and 21 for THUMOS-14.
To train IIN, we use the Adam~\cite{adam} optimizer with the initial learning rate of $10^{-4}$ for both datasets and set the batch size to 32.


\begin{table}[t]
\caption{\label{tab3}Ablation study of the effectiveness of our proposed components on TVSeries \cite{geest2016eccv}. CI and EE indicate additionally using the current information and early embedding input information, respectively. The best scores are marked in \textbf{bold}.}
\vspace{-0.3cm}
\centering
\begin{tabular}{C{5.5cm}|C{2.cm}}\hline
Method & mcAP (\%) \\ \hline\hline
RNN-Simple & 79.9\\
RNN-LSTM & 80.9 \\
RNN-GRU (Baseline) & 81.3 \\ \cdashline{1-2}
Baseline+CI & 83.4 \\
Baseline+CI+EE (IDN) & \bf{84.7} \\ \hline
\end{tabular}
\centering
\end{table}

\begin{table}[t]
\caption{\label{tab4}Ablation study of the effectiveness of our proposed components on THUMOS-14 \cite{jiang2015url}. CI and EE indicate additionally using the current information and early embedding input information, respectively. The best scores are marked in \textbf{bold}.}
\vspace{-0.3cm}
\centering
\begin{tabular}{C{5.5cm}|C{2.cm}}\hline 
Method& mAP (\%) \\ \hline\hline
RNN-Simple & 45.5 \\
RNN-LSTM & 46.3 \\ 
RNN-GRU (Baseline) & 46.7 \\ \cdashline{1-2}
Baseline+CI & 48.6 \\
Baseline+CI+EE (IDN) & \bf{50.0} \\ \hline
\end{tabular}
\centering
\end{table}
\begin{figure}[t]
\centering{\includegraphics[width=.97\linewidth]{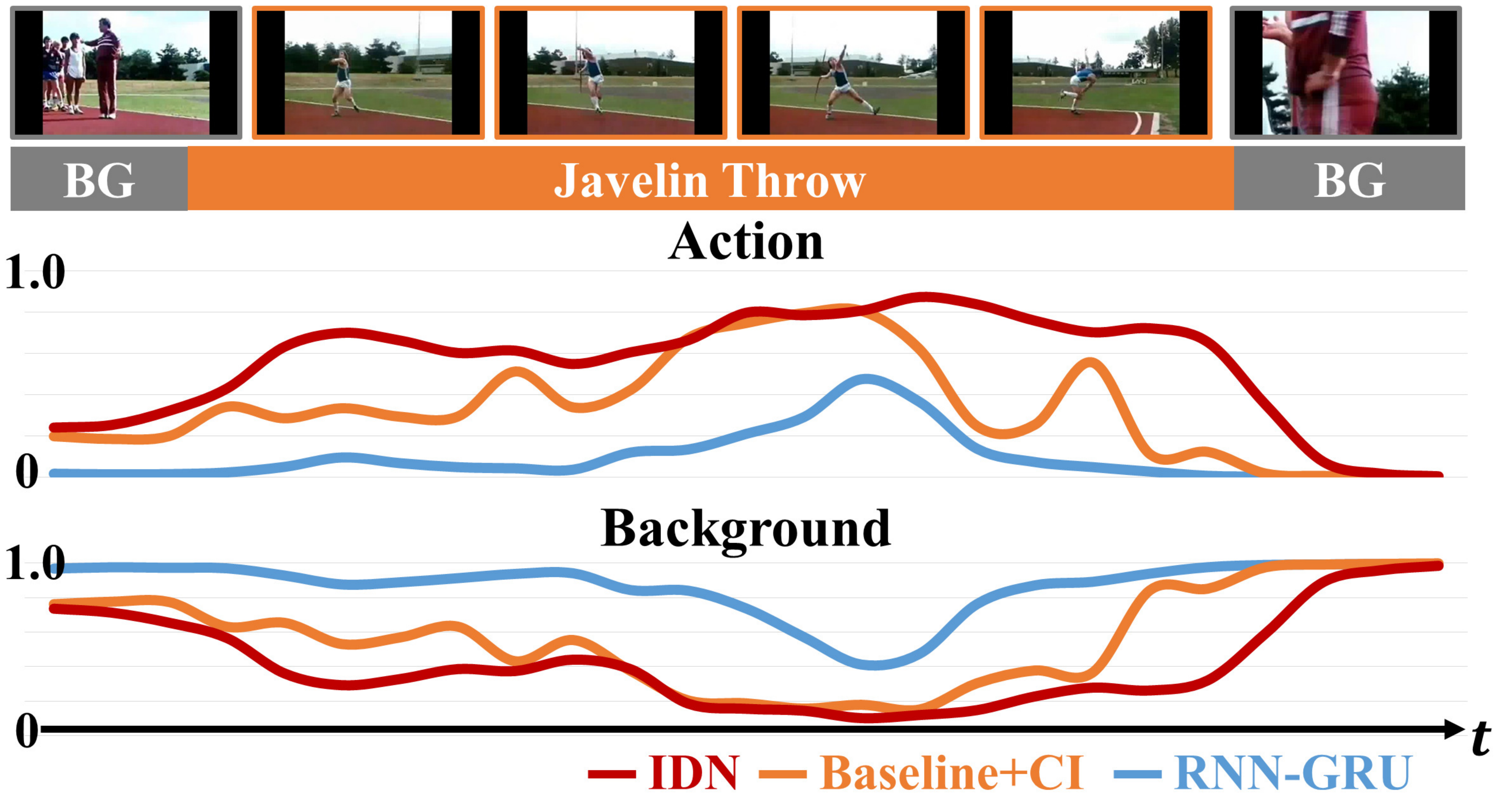}}
\centering{\caption{\label{fig4}Qualitative comparisons on predicted and GT probabilities for action (top) and background (bottom).}
}
\end{figure}


\section{Results and Analysis}
\subsection{Online Action Detection}
\subsubsection{Ablation Study} \label{sec:oad-ablation}
We evaluate RNNs with the simple unit, LSTM \cite{hochreiter1997nc}, and GRU \cite{cho2014emnlp}.
We name these networks RNN-Simple, RNN-LSTM, and RNN-GRU, respectively.
Although many methods \cite{geest2016eccv,gao22017bmvc,xu2019iccv} report the performances of these networks as baselines, we evaluate them in our setting to clearly confirm the effectiveness of our IDU.

In addition, we individually add IDU components to GRU as a baseline for analyzing their effectiveness:\\
\noindent \textbf{Baseline+CI:} We add a mechanism using current information to GRU in computing reset and update gates. 
Specifically, we replace Eq.~(\ref{eq:gru1}) for $r_t$ with
\begin{eqnarray}
r_t = \sigma (\textbf{W}_{hr}h_{t-1} + \textbf{W}_{x_0r}x_0)
\end{eqnarray}
and Eq.~(\ref{eq:gru3}) for $z_t$ with
\begin{eqnarray}
z_t = \sigma (\textbf{W}_{x_tz}x_t + \textbf{W}_{x_0z}x_0),
\end{eqnarray}
where $\textbf{W}_{hr}$, $\textbf{W}_{x_0r}$, $\textbf{W}_{x_tz}$, and $\textbf{W}_{x_0z}$ are trainable parameters.
We construct a recurrent network with this modified unit.\\
\textbf{Baseline+CI+EE (IDN):} We incorporate our main components, a mechanism utilizing current information and an early embedding module, into GRU, which is our IDU.
These components enable reset and update gates to effectively model the relationship between an ongoing action and input information at every time step.
Specifically, Eq.~(\ref{eq12}) and Eq.~(\ref{eq14}) are substituted for Eq.~(\ref{eq:gru1}) and Eq.~(\ref{eq:gru3}), respectively.
We design a recurrent network with our IDU, which is the proposed IDN.

\begin{figure}[t]
\centering{\includegraphics[width=.97\linewidth]{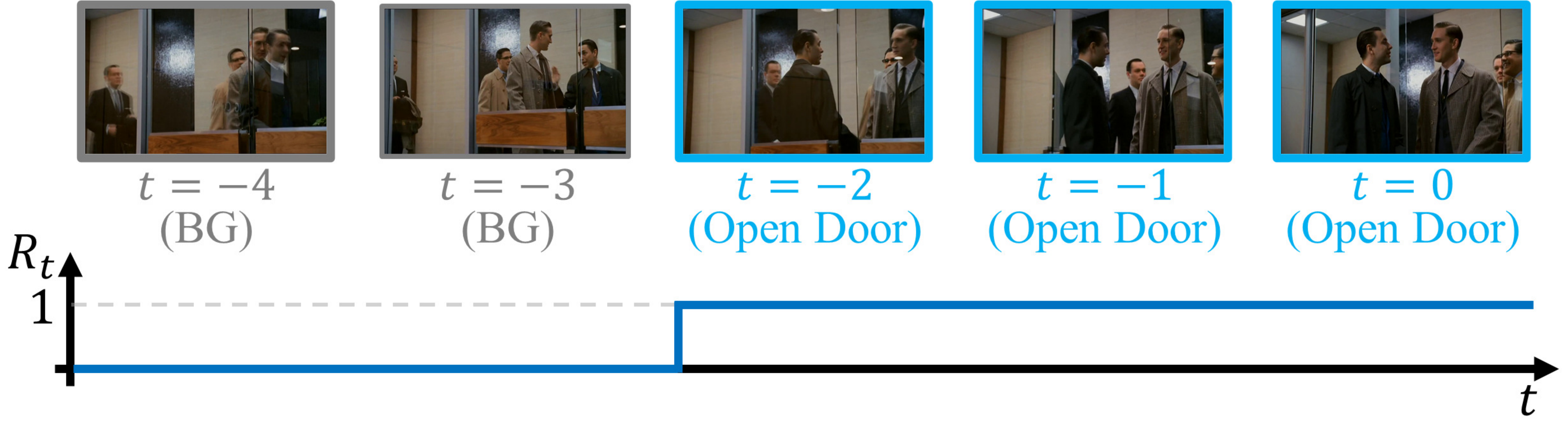}}
\centering{\caption{\label{fig:Rt}Example of relevance scores $R_t$ of input chunks.}}
\end{figure}
\begin{figure}[t]
\begin{minipage}[t]{0.99\linewidth}
\begin{center}
\includegraphics[width=.98\linewidth]{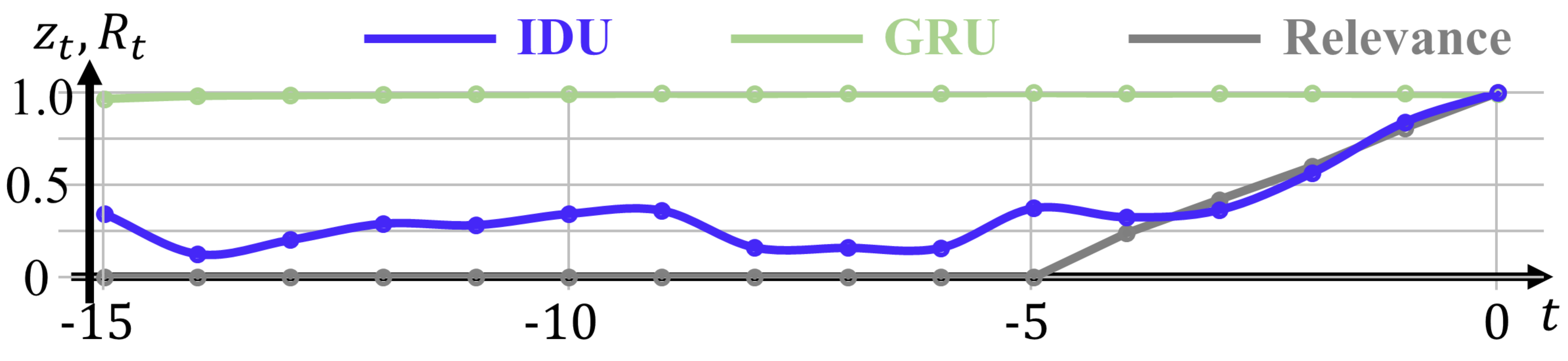}
\end{center}
\vspace{-0.3cm}
\footnotesize{(a) On the input sequences containing from one to five relevant chunks (i.e., from $t=0$ to $t=-4$). }
\end{minipage}
\hfill \vspace{0.1cm}
\begin{minipage}[t]{0.99\linewidth}
\begin{center}
\includegraphics[width=.98\linewidth]{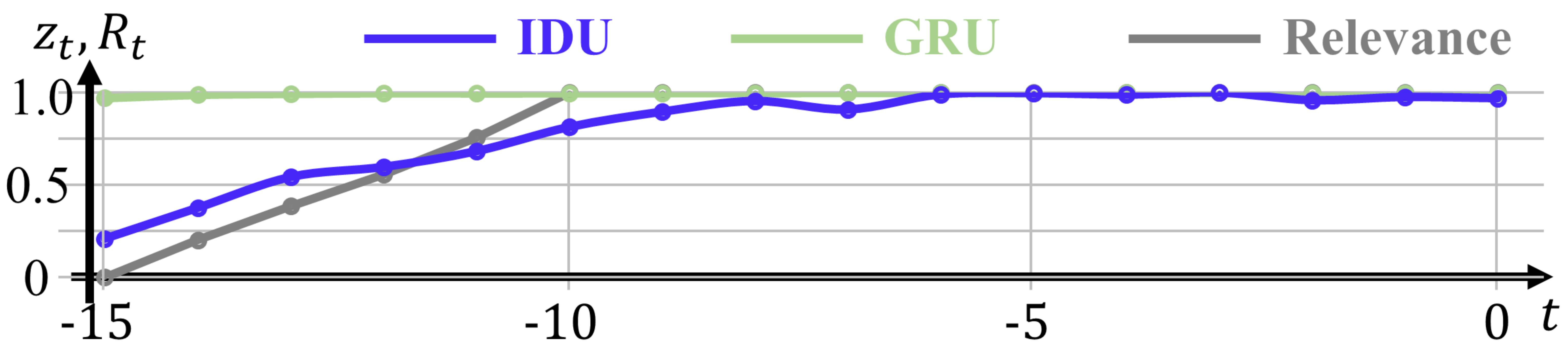}
\end{center}
\vspace{-0.3cm}
\footnotesize{(b) On the input sequences containing from 11 to 15 relevant chunks (i.e., from $t=-10$ to $t=-14$).}
\end{minipage}
\hfill \vspace{0.2cm}
\caption{\label{fig:zt}Comparison between the update gate $z_t$ values of our IDU and GRU \cite{cho2014emnlp}. Update gate values are measured on the input sequences containing (a) from one to five relevant chunks and (b) from 11 to 15 relevant chunks.}
\end{figure}

In Table \ref{tab3}, we report the performances of five networks on the TVSeries dataset \cite{geest2016eccv}.
Among RNN-Simple, RNN-LSTM, and RNN-GRU, RNN-GRU results in the highest mcAP of 81.3\%.
By comparing RNN-GRU (Baseline) with Baseline+CI, we first analyze the effect of using $x_0$ in calculating reset and update gates.
This component enables the gates to decide whether input information at each time is relevant to a current action.
As a result, Baseline-CI achieves the performance gain of 2.1\% mcAP, which demonstrates the effectiveness of using $x_0$.
Next, we observe that adding the early embedding module improves the performance by 1.3\% mcAP from the comparison between Baseline+CI and Baseline+CI+EE (IDN).
Note that our IDN achieves mcAP of 84.7\% with a performance gain of 3.4\% mcAP compared with Baseline.
We conduct the same experiment on the THUMOS-14 dataset \cite{jiang2015url} to confirm the generality of the proposed components.
We obtain performance gains as individually incorporating the proposed components into GRU (see Table \ref{tab4}), where our IDN achieves improvements of 3.3\% mAP compared to Baseline.
These results successfully demonstrate the effectiveness and generality of our components.

Figure \ref{fig4} shows qualitative comparisons on predicted and GT probabilities, where our IDN achieves the best results on both action and background frames.
To confirm the effect of our components, we compare the values of the update gates $z_t$ between our IDU and GRU.
For a reference, we introduce the relevance score $R_t$ of each chunk regarding a current action.
Specifically, we set the scores of input chunks representing the current action as 1, otherwise 0 (see Fig. \ref{fig:Rt}).
Note that the update gate controls how much information from the input will carry over to the hidden state.
Therefore, the update gate should drop the irrelevant information and pass over the relevant information related to the current action.
In Fig. \ref{fig:zt}, we plot the $z_t$ values of IDU and GRU and relevance scores against each time step.
On the input sequences containing from one to five relevant chunks, the $z_t$ values of GRU are very high at all time steps.
In contrast, our IDU successfully learns the $z_t$ values following the relevance scores (see Fig. \ref{fig:zt} (a)).
We also plot the average $z_t$ values on the input sequences including from 11 to 15 relevant chunks in Fig. \ref{fig:zt} (b), where our IDU yields the $z_t$ values similar to the relevance scores.
These results demonstrate that our IDU effectively models the relevance of input information to the ongoing action.

Compared to GRU, IDU has additional weights ${\bf{W}}_{xe}\in \mathbb{R}^{d_x \times 512}$ and ${\bf{W}}_{ep}\in \mathbb{R}^{512 \times (K+1)}$ in the early embedding module.
Our early embedding module reduces the dimensions of $x_t$, $x_0 \in \mathbb{R}^{d_x \times 512}$, which makes the parameters (i.e., ${\bf{W}}_{x_0r}$, ${\bf{W}}_{x_tz} \in \mathbb{R}^{512 \times 512}$) in IDU less than the parameters (i.e., ${\bf{W}}_{xr}$, ${\bf{W}}_{xz}\in \mathbb{R}^{d_x \times 512}$) in GRU.
The other weights have the same number of parameters in IDU and GRU.
As a result, the number of parameters in IDU is 57.4\% of that in GRU with $d_x=3072$ and $K=20$. 
Also, FLOPs of IDU is 57.3\% of that of GRU with the same hyper-parameters.

\begin{figure*}[t]
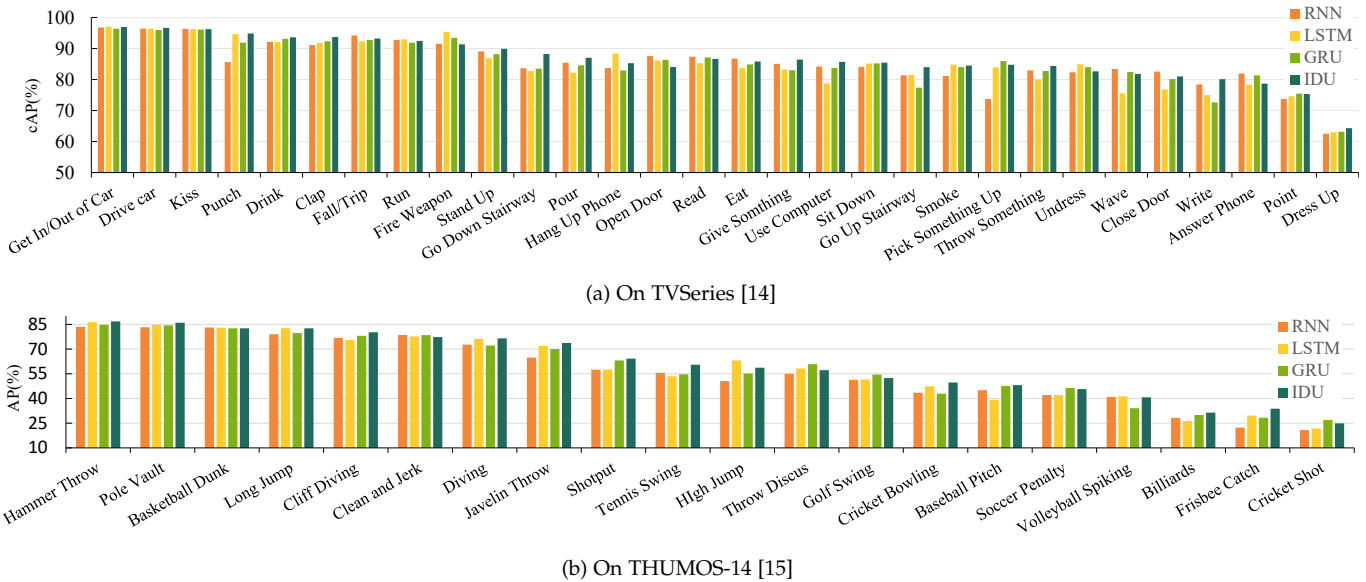

\centering
{\centering{\includegraphics[width=.99\linewidth]{./figures/class_tvseries.pdf}}}
\centering{\footnotesize{(a) On TVSeries \cite{geest2016eccv}}}
\centering{\includegraphics[width=.99\linewidth]{./figures/class_thumos14.pdf}} 
\centering{\footnotesize{(b) On THUMOS-14 \cite{jiang2015url}}}
\vspace{-0.2cm}
\centering{\caption{\label{fig7}Performance comparison for each class on (a)TVSeries~\cite{geest2016eccv} and (b)THUMOS-14~\cite{jiang2015url}. Action classes are sorted in descending order of IDU performance.}}
\end{figure*}
\subsubsection{Discussion on performance comparison of different action classes}



In Fig.~\ref{fig7}, we report a cAP value on TVSeries~\cite{geest2016eccv} and an AP value on THUMOS-14~\cite{jiang2015url} for each class, which is generated by three baseline (i.e., RNN, LSTM, and GRU) and our proposed IDU.
These four models have similar performance tendency actions that the higher performances are achieved for actions that have large motions of actors such as `Drive Car' and `Get In/Out of Car' on TVSeries, and ‘Hammer Throw’ and ‘Pole Vault’ on THUMOS-14.
On the other hand, the lower performances are observed for actions, which have small motions of small objects without large motions of actors such as `Point' and `Dress Up' on TVSeries, and `Cricket Shot' and `Billiards' on THUMOS-14.

\begin{table}[t!]
\centering\caption{\label{tab5}Performance comparison on TVSeries \cite{geest2016eccv}. IDN, TRN \cite{xu2019iccv}, RED \cite{gao22017bmvc}, and ED \cite{gao22017bmvc} use same two-stream features for the Two-Stream input. The best scores are marked in \textbf{bold}.}
\vspace{-0.3cm}
\centering
\begin{tabular}{C{2.5cm}|C{3.cm}|C{1.5cm}}\hline 
Input & Method & mcAP (\%) \\ \hline\hline
\multirow{5}{*}{RGB}&LRCN \cite{donahue2015cvpr} & 64.1 \\
&RED \cite{gao22017bmvc}		 & 71.2 \\
&2S-FN \cite{geest2018wacv} & 72.4 \\
&TRN \cite{xu2019iccv} 		 & 75.4 \\
&IDN      		 & \bf{76.6} \\ \hline
\multirow{2}{*}{Flow}&FV-SVM \cite{geest2016eccv} & 74.3 \\ 
&IDN 		 & \bf{80.3} \\ \hline
\multirow{4}{*}{\shortstack{Two-Stream}}&RED \cite{gao22017bmvc}		 & 79.2 \\
&TRN \cite{xu2019iccv} 		 & 83.7 \\
&IDN      		 & \bf{84.7} \\ \cdashline{2-3}
&IDN-Kinetics      		 & \bf{86.1} \\ \hline
\end{tabular}
\end{table}

\subsubsection{Performance Comparison}
\label{sec:oad_per}
In this section, we compare our IDN with state-of-the-art methods on TVSeries \cite{geest2016eccv} and THUMOS-14 \cite{jiang2015url} datasets.
We use three types of input, including RGB, Flow, and Two-Stream.
As the input of our IDU, we take only appearance features for the RGB input and motion features for the Flow input.
IDN, TRN \cite{xu2019iccv}, RED \cite{gao22017bmvc}, and ED \cite{gao22017bmvc} use the same two-stream features for the Two-Stream input, which allows a fair comparison.
We also employ another feature extractor, the TSN model \cite{wang2016eccv} pretrained on the Kinetics dataset \cite{carreira2017cvpr}.
We name our IDN with this feature extractor IDN-Kinetics.
\begin{table}[t!]
\caption{\label{tab6}Performance comparison on THUMOS-14 \cite{jiang2015url}. IDN, TRN \cite{xu2019iccv}, RED \cite{gao22017bmvc}, and ED \cite{gao22017bmvc} use same two-stream features. The best scores are marked in \textbf{bold}.}
\vspace{-0.3cm}
\centering
\begin{tabular}{C{2.5cm}|C{3.cm}|C{1.5cm}}\hline 
Setting & Method & mAP (\%) \\ \hline\hline
\multirow{5}{*}{Offline} & CNN \cite{simonyan2015iclr} & 34.7 \\
 & CNN \cite{simonyan2014nips} & 36.2 \\
 & LRCN \cite{donahue2015cvpr}  & 39.3 \\
 & MultiLSTM \cite{yeung2018ijcv} & 41.3 \\
 & CDC \cite{shou2017cvpr} & 44.4 \\ \hline
\multirow{4}{*}{Online}  & RED \cite{gao22017bmvc} & 45.3 \\
 & TRN \cite{xu2019iccv} 		 & 47.2 \\
 & IDN		      		 & \bf{50.0} \\ \cdashline{2-3}
 & IDN-Kinetics      		 & \bf{60.3} \\ \hline
\end{tabular}
\centering
\end{table}
\begin{table*}[t]
\centering
\caption{\label{tab7}Performance comparison for different portions of  actions on TVSeries \cite{geest2016eccv} in terms of mcAP (\%). The corresponding portions of actions are only used to compute mcAP after detecting current actions on all frames in an online manner. The best scores are marked in \textbf{bold}.}
\vspace{-0.3cm}
\begin{tabular}{C{4.5cm} | C{.8cm} C{.8cm} C{.8cm} C{.8cm} C{.8cm} C{.8cm} C{.8cm} C{.8cm} C{.8cm} C{.8cm}}\hline 

\multirow{3}{*}{Method} & \multicolumn{10}{c}{Portion of action} \\ \cline{2-11}
 & 0\%-10\% & 10\%-20\% & 20\%-30\% & 30\%-40\% & 40\%-50\% & 50\%-60\% & 60\%-70\% & 70\%-80\% & 80\%-90\% & 90\%-100\% \\ \hline\hline
\multicolumn{1}{c|}{CNN \cite{geest2016eccv}} & 61.0 & 61.0 & 61.2 & 61.1 & 61.2 & 61.2 & 61.3 & 61.5 & 61.4 & 61.5 \\
\multicolumn{1}{c|}{LSTM \cite{geest2016eccv}} & 63.3 & 64.5 & 64.5 & 64.3 & 65.0 & 64.7 & 64.4 & 64.4 & 64.4 & 64.3 \\
\multicolumn{1}{c|}{FV-SVM \cite{geest2016eccv}} & 67.0 & 68.4 & 69.9 & 71.3 & 73.0 & 74.0 & 75.0 & 75.4 & 76.5 & 76.8 \\
\multicolumn{1}{c|}{TRN \cite{xu2019iccv}} & 78.8 & 79.6 & 80.4 & 81.0 & 81.6 & 81.9 & 82.3 & 82.7 & 82.9 & 83.3 \\
\multicolumn{1}{c|}{IDN}  & 80.6 & 81.1 & 81.9 & 82.3 & 82.6 & 82.8 & 82.6 & 82.9 & 83.0 & 83.9 \\ \cdashline{1-11}
\multicolumn{1}{c|}{IDN-Kinetics}  & \bf{81.7} & \bf{81.9} & \bf{83.1} & \bf{82.9} & \bf{83.2} & \bf{83.2} & \bf{83.2} & \bf{83.0} & \bf{83.3} & \bf{86.6} \\ \hline
\end{tabular}
\end{table*}

\begin{figure*}[t!]
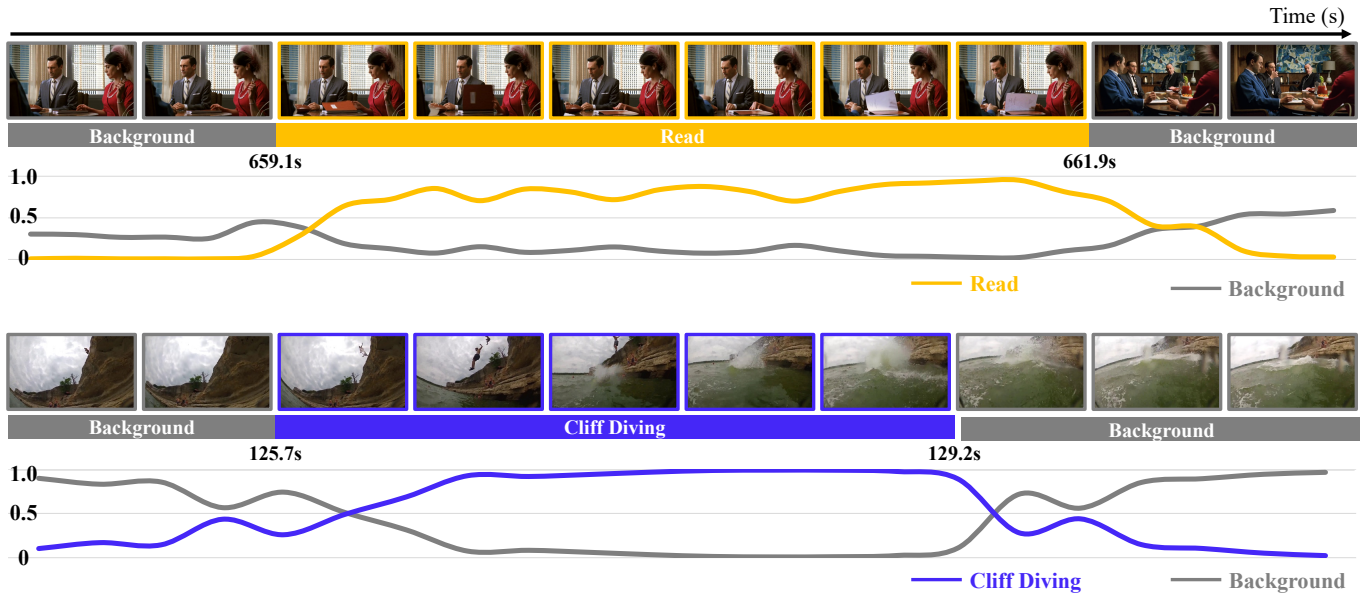

\begin{minipage}[t]{0.99\linewidth}
\begin{center}
\includegraphics[width=\linewidth]{./figures/fig11-a_revision.pdf}
\end{center}
\centering
\end{minipage}
\hfill 
\vspace{0.3cm}
\begin{minipage}[t]{0.99\linewidth}
\begin{center}
\includegraphics[width=\linewidth]{./figures/fig11-b_revision.pdf}
\end{center}
\vspace{-0.1cm}
\centering
\end{minipage}
\hfill \vspace{-0.1cm}
\caption{\label{fig:idn_result}Qualitative evaluation of IDN on TVSeries \cite{geest2016eccv} (upper) and THUMOS-14 \cite{jiang2015url} (lower). Each result shows frames, ground truth, and estimated probabilities.}
\end{figure*}
We report the results on TVSeries in Table \ref{tab5}.
Our IDN significantly outperforms state-of-the-art methods on all types of input, where IDN achieves 76.6\% mcAP on the RGB input, 80.3\% mcAP on the Flow input, and 84.1\% mcAP on the Two-Stream input.
Furthermore, IDN-Kinetics achieves the best performance of 86.1\% mcAP.
Note that IDN effectively reduces wrong detection results occurred from the irrelevant information by discriminating the relevant information.
However, 2S-FN, RED, and TRN accumulate the input information without considering its relevance to an ongoing action.
In addition, our IDN yields better performance than TRN \cite{xu2019iccv} although IDN takes shorter temporal information than IDN (i.e., 16 chunks vs. 64 chunks).

In Table \ref{tab6}, we compare performances between our IDN and state-of-the-art approaches for online and offline action detection.
The compared offline action detection methods perform frame-level prediction.
As a result, both IDN and IDN-Kinetics outperform all methods by a large margin.

In online action detection, it is important to identify actions as early as possible.
To compare this ability, we measure the mcAP values for every 10\% portion of actions on TVSeries.
Table \ref{tab7} shows the comparison results among IDN, IDN-Kinetics, and previous methods, where our methods achieve state-of-the-art performance at every time interval.
This demonstrates the superiority of our IDU in identifying actions at early stages as well as all stages.
\begin{figure}[t!]
\centering{\includegraphics[width=.97\linewidth]{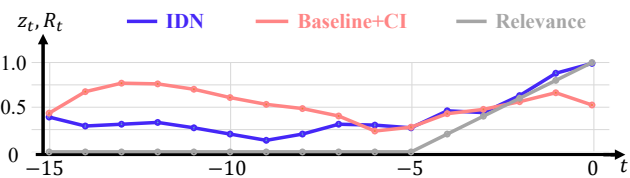}}
\centering{\caption{\label{fig_rt}Comparison between the update gate $z_t$ values of IDN with and without the early embedding module. Baseline+CI is IDN without the early embedding module.}
}
\end{figure}

\subsubsection{Qualitative Evaluation}
For qualitative evaluation, we visualize our results on TVSeries \cite{geest2016eccv} and THUMOS-14~\cite{jiang2015url} in Fig.~\ref{fig:idn_result}.
The results on the TVSeries dataset show high probabilities on the true action label and reliable start and end time points.
Note that identifying actions at the early stage is very challenging in this scene because only subtle changes happen.
On THUMOS-14, our IDN successfully identifies ongoing actions by yielding the contrasting probabilities between true action and background labels.

\begin{table}[t!]
\caption{\label{tab:ee_loss}Ablation study of the classification loss $L_{ee}$ and the contrastive loss $L_{ct}$ of the proposed early embedding module on THUMOS-14~\cite{jiang2015url} and TVSeries~\cite{geest2016eccv}. The best scores are marked in \textbf{bold}.}
\centering
\vspace{-0.3cm}
\begin{tabular}{C{1.2cm}C{1.2cm}|C{2.1cm}C{2.1cm}}
\hline
\multicolumn{2}{c|}{Losses}                           & THUMOS-14 & TVseries \\
$L_{ee}$                  & $L_{ct}$                  & mAP(\%)   & mcAP(\%) \\ \hline \hline
                          &                           & 48.6      & 83.4    \\
\checkmark &                           & 49.3     & 84.0    \\
                          & \checkmark & 49.4     & 84.2    \\
\checkmark & \checkmark & \bf{50.0}      & \bf{84.7}     \\ \hline
\end{tabular}
\centering
\end{table}
\subsubsection{Discussion on the early embedding module} \label{sec:ee}
In this section, we explain the effect of the early embedding module in detail.
With the action class loss $L_{ee}$ and the feature distance loss $L_{ct}$, the early embedding module forces $x_{t}^{e}$ to represent specific actions.
By doing so, the early embedding module allows the update and forget gate to focus on accumulating relevant information to current action.
To validate this, in Figure~\ref{fig_rt}, we compare the update gate $z_t$ values of IDN with and without the early embedding, which are denoted as IDN and Baseline+CI, respectively. 
We observe that Baseline+CI has a high $z_t$ value even at a step with a small relevance score (i.e., from 15 to 10 chunks). 
On the other hand, IDN learns the $z_t$ values following the relevance score $R_t$.
This observation demonstrates that using the early embedding module not only reduces the number of parameters but also encourages the reset and update gate to discriminate the relevant information to current action.”.

We conduct ablation experiments about the two losses $L_{ee}$ and $L_{ct}$ in the proposed early embedding module.
The results are summarized in Table~\ref{tab:ee_loss}.
IDN with only $L_{ee}$ obtains 49.3\% mAP and 84.0\% mcAP on THUMOS-14 and TVSeries, respectively.
IDN with only $L_{ct}$ achieves 49.4\% mAP on THUMOS-14 and 84.2\% mcAP on TVSeries.
By training the model with both $L_{ee}$ and $L_{ct}$, we obtain the best performance of 50.0\% mAP and 84.7\% mcAP on THUMOS-14 and TVSereis, respectively.
\begin{table}[t!]
\centering
\caption{\label{tab:ab_aa1}Ablation study of the effectiveness of our proposed IIU on  TVSeries~\cite{geest2016eccv}. The best and the second best scores are marked in \textbf{bold}.}
\vspace{-0.3cm}
\centering
{\begin{tabular}{C{3cm} | C{2cm} C{2cm}}
\hline
\multirow{2}{*}{Method}           & \multicolumn{2}{c}{\begin{tabular}[c]{@{}c@{}}Time predicted into the future \\ (seconds)\end{tabular}} \\ \cline{2-3}
                   
                                  & 1.0s               & 2.0s               \\ \hline \hline
\multicolumn{1}{c|}{Simple RNN}         & 69.1               & 66.5               \\
\multicolumn{1}{c|}{LSTM}         & 69.6               & 65.1               \\
\multicolumn{1}{c|}{IDU}         & 72.7               & 70.3               \\
\multicolumn{1}{c|}{GRU}          & 73.9               & 71.2               \\
\multicolumn{1}{c|}{IIU}          & \textbf{76.7}               & \textbf{74.3} \\ \cdashline{1-3}
\multicolumn{1}{c|}{IIU-Oracle} & \textbf{97.2} & \textbf{85.4} \\  \hline
\end{tabular}
}
\end{table}


\begin{table}[t!]
\centering
\caption{\label{tab:ab_aa2}Ablation study of the effectiveness of our proposed IIU on THUMOS-14~\cite{jiang2015url}. The best and the second best scores are marked in \textbf{bold}.}
\vspace{-0.3cm}
\centering
{\begin{tabular}{C{3cm} | C{2cm} C{2cm}}
\hline
\multirow{2}{*}{Method}           & \multicolumn{2}{c}{\begin{tabular}[c]{@{}c@{}}Time predicted into the future \\ (seconds)\end{tabular}} \\ \cline{2-3} 
                                  & 1.0s             & 2.0s               \\ \hline \hline
\multicolumn{1}{c|}{Simple RNN}         & 47.1               &37.7                \\
\multicolumn{1}{c|}{LSTM}         & 47.5             & 39.7               \\
\multicolumn{1}{c|}{IDU}         & 47.8              & 40.4               \\
\multicolumn{1}{c|}{GRU}          & 48.4               & 40.0               \\
\multicolumn{1}{c|}{IIU}          & \textbf{52.0}               & \textbf{47.2}  \\
\cdashline{1-3}
\multicolumn{1}{c|}{IIU-Oracle} & \textbf{88.5} & \textbf{73.7} \\  \hline
\end{tabular}
}
\end{table}

\subsection{Action Anticipation}
\subsubsection{Ablation Study} \label{sec:iiu-ablation}
To show the effectiveness of our IIU, we conduct ablation studies by strategically replacing IIU with conventional recurrent units  (i.e., simple RNN, LSTM \cite{hochreiter1997nc} and GRU \cite{cho2014emnlp}) and IDU in IIN. 
For the experiments, we use the concatenated feature $[x_t \: || \: g_t]$ as an input of recurrent units. 
Also, to demonstrate the utility of action labels on action anticipation, we experiment with the network named IIN-Oracle. 
IIN-Oracle has the same architecture as IIN but uses ground-truth action labels instead of pseudo action labels. 
In these experiments, we evaluate the performances on action anticipation after 1 and 2 seconds from the last observation.

In Table~\ref{tab:ab_aa1} and~\ref{tab:ab_aa2}, we summarize the results of four models on TVSeries~\cite{geest2016eccv} and THUMOS-14~\cite{jiang2015url} datasets, respectively. 
By comparing a simple recurrent unit (i.e., RNN), LSTM and GRU with IIU, we demonstrate the ability of our IIU to integrate different modality features into enriched features. 
Consequentially, compared to the network with GRU on TVSeries, the network with IIU (i.e., IIN) achieves 76.7\% mcAP and 74.3\% mcAP with performance gains of 2.8\% mcAP and 3.1\% mcAP for predicting future action after 1 and 2 seconds, respectively. 
Similarly, on THUMOS-14, our IIN achieves performance improvements by 3.6\% mAP and 7.2\% mAP than the network with GRU for predictions at 1 and 2 seconds, respectively.
Also, our IIU achieves higher performances than IDU on both TVSeries and THUMOS-14 datasets. 
From these results, we show that, on action anticipation, contextualizing observed actions is more effective than accumulating current information. 
Comparing IIN-Oracle to IIN, using ground truth action classes significantly improves the performance of action anticipation. 
This result demonstrates that action labels are prominent cues on action anticipation.
\begin{figure}[t]
\centering{\includegraphics[width=.97\linewidth]{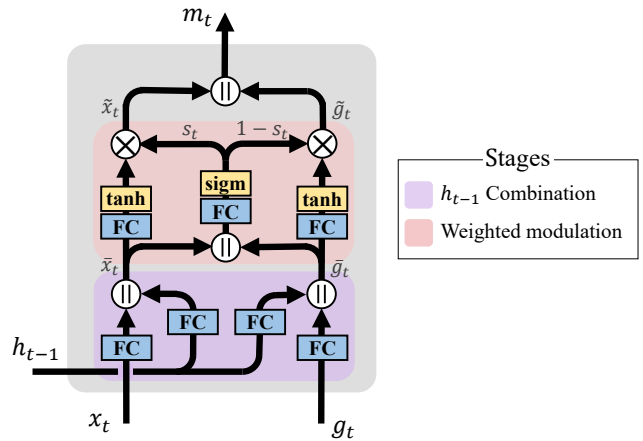}}
\centering{\caption{\label{fig:integration_module} The integration module is divided into two stages: $h_{t-1}$ combination stage and weighted modulation stage.}
}
\end{figure}


\begin{table}[t]
\caption{\label{tab:int_module}Ablation study of two components in the integration module on THMOS-14~\cite{jiang2015url}. `$h_{t-1}$ Comb.` and `W. Mod.` indicate $h_{t-1}$ combination stage and weighted modulation stage, respectively. The best scores are marked in \textbf{bold}.}
\centering
\vspace{-0.3cm}
\begin{tabular}{C{1.1cm}C{1.1cm}|C{2.0cm}C{2.0cm}}
\hline
\multirow{2}{*}{\begin{tabular}[c]{@{}c@{}}$h_{t-1}$\\  Comb.\end{tabular}} & \multirow{2}{*}{W. Mod.} &  \multicolumn{2}{c}{\begin{tabular}[c]{@{}c@{}}Time predicted into the future\\ (seconds)\end{tabular}} \\ \cline{3-4} 
  & & 1.0s & 2.0s \\ \hline \hline
 &  &   48.8 &43.0  \\ 
 & \checkmark &  51.2& 44.7 \\ 
 \checkmark &\checkmark  & \textbf{52.0}  & \textbf{47.2}  \\ \hline
\end{tabular}
\centering
\end{table}


Compared to GRU, our IIU has additional weights for modality integration. 
The number of parameters in IIU is 2.3 times more than that in GRU.
Also, IIU has 876.94M FLOPs, which is 2.1 times more than that of GRU. 
From an input video $\mathbf{V}$ and a pseudo action label $\mathbf{A}$ to an output $\mathbf{q}$, our IIN takes 99.4ms for inference. 
Since IIN need pseudo labels generated by IDN, it takes 196.5ms from an input V for action anticipation

\begin{table*}[t!]
\caption{\label{tab8}Performance comparison on TVSeries~\cite{geest2016eccv} in terms of mcAP(\%).
The best and second best scores are marked in \textbf{bold}.
 }
\vspace{-0.3cm}
\centering
{\begin{tabular}{C{4cm} | C{1.1cm} C{1.1cm} C{1.1cm} C{1.1cm} C{1.1cm} C{1.1cm} C{1.1cm} C{1.1cm} | C{1.1cm}}
\hline
\multirow{2}{*}{Method} & \multicolumn{8}{c|}{Time predicted into the future (seconds)}                                                                  & \multirow{2}{*}{Avg.}\\ \cline{2-9}
                        & 0.25s         & 0.5s          & 0.75s         & 1.0s          & 1.25s         & 1.5s          & 1.75s         & 2.0s          &                  \\ \hline \hline
\multicolumn{1}{c|}{ED~\cite{gao22017bmvc}}           & 78.5          & 78.0          & 76.3          & 74.6          & 73.7          & 72.7          & 71.7          & 71.0          & 74.5                 \\
\multicolumn{1}{c|}{RED~\cite{gao22017bmvc}}          & 79.2          & 78.7          & 77.1          & 75.5          & 74.2          & 73.0          & 72.0          & 71.2          & 75.1                 \\
\multicolumn{1}{c|}{TRN~\cite{xu2019iccv}}          & \textbf{79.9}  &\textbf{78.4} & 77.1          & 75.9          & 74.9          & 73.9          & 73.0          & 72.3          & 75.7                 \\ 
\multicolumn{1}{c|}{IIN}          & 77.8 & 77.6 & \textbf{77.2} & \textbf{76.7} & \textbf{76.4} & \textbf{75.6} & \textbf{75.3} & \textbf{74.3} & \textbf{76.4} \\ \cdashline{1-10}
\multicolumn{1}{c|}{IIN-Kinetics}& \textbf{80.0} & \textbf{79.7} & \textbf{79.2} & \textbf{78.6} & \textbf{78.2} & \textbf{77.8} & \textbf{77.3} & \textbf{76.6} & \textbf{78.4}       \\ \hline
\end{tabular}}
\end{table*}

\begin{table*}[t!]
\caption{\label{tab9}Performance comparison on THUMOS-14 \cite{jiang2015url} in terms of mAP(\%).
The best and second best scores are marked in \textbf{bold}.
}
\vspace{-0.3cm}
\centering
\begin{tabular}{C{4cm} | C{1.1cm} C{1.1cm} C{1.1cm} C{1.1cm} C{1.1cm} C{1.1cm} C{1.1cm} C{1.1cm} | C{1.1cm}}
\hline
\multirow{2}{*}{Method}           & \multicolumn{8}{c|}{Time predicted into the future (seconds)}                                                                  & \multirow{2}{*}{Avg.} \\ \cline{2-9}
                                  & 0.25s         & 0.5s          & 0.75s         & 1.0s          & 1.25s         & 1.5s          & 1.75s         & 2.0s          &                  \\ \hline \hline
\multicolumn{1}{c|}{ED~\cite{gao22017bmvc}}           & 43.8          & 40.9          & 38.7          & 36.8          & 34.6          & 33.9          & 32.5          & 31.6          & 36.6                 \\
\multicolumn{1}{c|}{RED~\cite{gao22017bmvc}}          & 45.3          & 42.1          & 39.6          & 37.5          & 35.8          & 34.4          & 33.2          & 32.1          & 37.5                 \\
\multicolumn{1}{c|}{TRN~\cite{xu2019iccv}}          & 45.1 & 42.4 & 40.7          & 39.1          & 37.7          & 36.4          & 35.3          & 34.3          & 38.9                 \\ 
\multicolumn{1}{c|}{IIN}         & \textbf{54.3} & \textbf{53.8} & \textbf{53.0} & \textbf{52.0} & \textbf{50.9} & \textbf{49.8} & \textbf{48.6} & \textbf{47.2} & \textbf{51.2}        \\ \cdashline{1-10}
\multicolumn{1}{c|}{IIN-Kinetics} & \textbf{55.6} & \textbf{55.3} & \textbf{54.6} & \textbf{53.1} & \textbf{51.4} & \textbf{49.8} & \textbf{48.5} & \textbf{46.9} & \textbf{51.9}       \\ \hline
\end{tabular}
\end{table*}
\begin{figure*}[t]
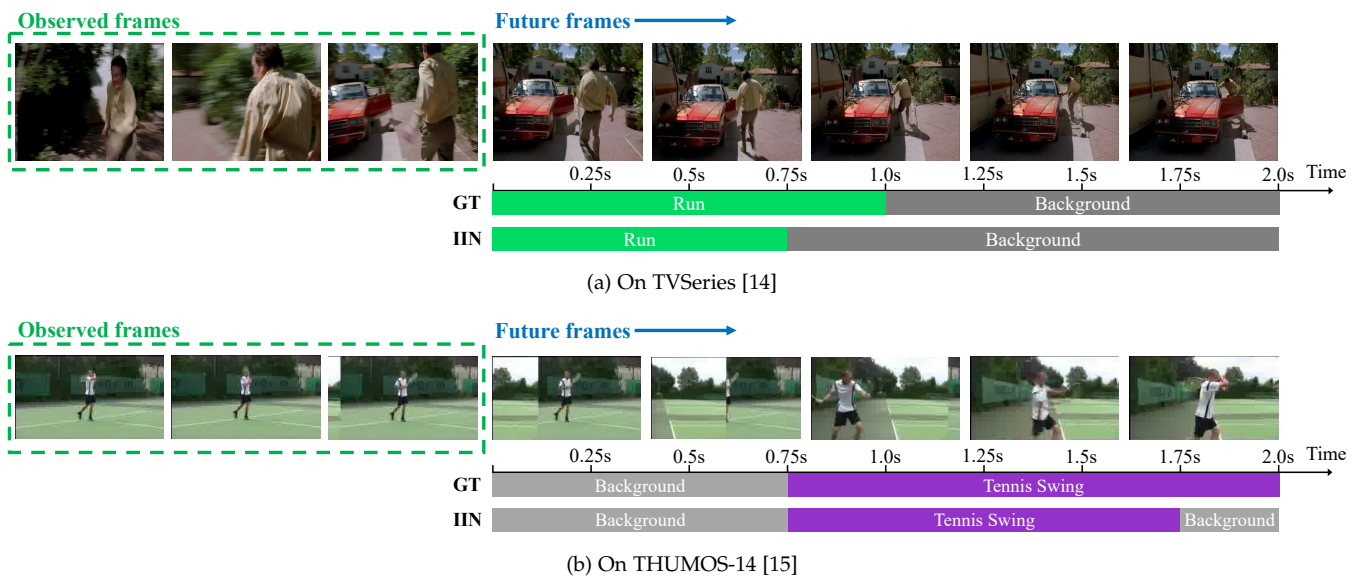

\begin{minipage}[t]{0.99\linewidth}
\begin{center}
\includegraphics[width=\linewidth]{./figures/fig13-a.pdf}
\centering{\footnotesize{(a) On TVSeries \cite{geest2016eccv}}}
\end{center}
\centering
\end{minipage}
\hfill 
\vspace{0.3cm}
\begin{minipage}[t]{0.99\linewidth}
\begin{center}
\includegraphics[width=\linewidth]{./figures/fig13-b.pdf}
\centering{\footnotesize{(b) On THUMOS-14 \cite{jiang2015url}}}
\end{center}
\vspace{-0.1cm}
\centering
\end{minipage}
\hfill \vspace{-0.1cm}
\caption{\label{fig:aa}Qualitative evaluation of IIN for action anticipation (a)TVSeries~\cite{geest2016eccv} and (b)THUMOS-14~\cite{jiang2015url} datasets.}
\end{figure*}
\subsubsection{Analysis of Integration Module}
In this section, we validate the superiority of the proposed the integration module in IIU.
The integration module is divided into two stages:
1) First, previous hidden state features are combined with visual and action label features respectively to encourage to exploit historical action information for predicting unseen future actions.
2) Then, $s_t$ is estimated to assign weights to visual and action label features.
We denote these two steps as $h_{t-1}$ combination stage and weighted modulation stage, respectively (see Fig.~\ref{fig:integration_module}).
To show ability of each part, we conduct ablation experiments of two stages.
The experimental results are summarized in Table~\ref{tab:int_module}.
While achieving 51.2 \% mAP at 1 second that is lower than IIU by only 0.8\% mAP, IIU without the $h_{t-1}$ combination stage obtains a performance lower performance than IIU by 2.5\% mAP at 2 seconds.
From these results, we demonstrate that, in the $h_{t-1}$ combination stage, comprehensive understanding of observed actions is richly exploited enough to forecast actions at further future.
Also, the weighted modulation stage improves performances compared to IIU with only the concatenation stage by 2.4\% and 1.7\% mAP at 1 and 2 seconds, respectively.
These results show that the weighted modulation stage controls an integration of visual and action label information to extract meaningful action features for action anticipation.

\subsubsection{Performance Comparison}
We compare our IIN with three state-of-the-art methods, ED~\cite{gao22017bmvc}, RED~\cite{gao22017bmvc}, and TRN~\cite{xu2019iccv}, for action anticipation.
Similar to the experiments on IDN, we use the same two-stream features $\{x_t\}_{t=-T}^{0}$ to TRN, RED, and ED for a fair comparison.
Also, we conduct experiments with features of the TSN architecture~\cite{wang2016eccv} pretrained on the Kinetics dataset~\cite{carreira2017cvpr}.
We call this network IIN-Kinetics.

We summarize the results of IIN on TVSeries~\cite{geest2016eccv} in Table~\ref{tab8}.
Our IIN outperforms the state-of-the-art methods at most of the future times.
IIN achieves 76.7\% mcAP for actions after 1 second, $74.3\%$ mcAP for actions after 2 seconds, and $76.4\%$ mcAP on average.
Moreover, we obtain the best performance at all of the future times in IIN-Kinetics, which achieves $78.4\%$ mcAP on average.
Although IIN obtains lower mcAP than TRN at 0.25 and 0.5 prediction times, it achieves stronger performance improvement at longer prediction times.
Note that TVSeries is a very challenging dataset due to sudden scene changes and the appearances of multiple people in a scene.

\begin{figure}[t]
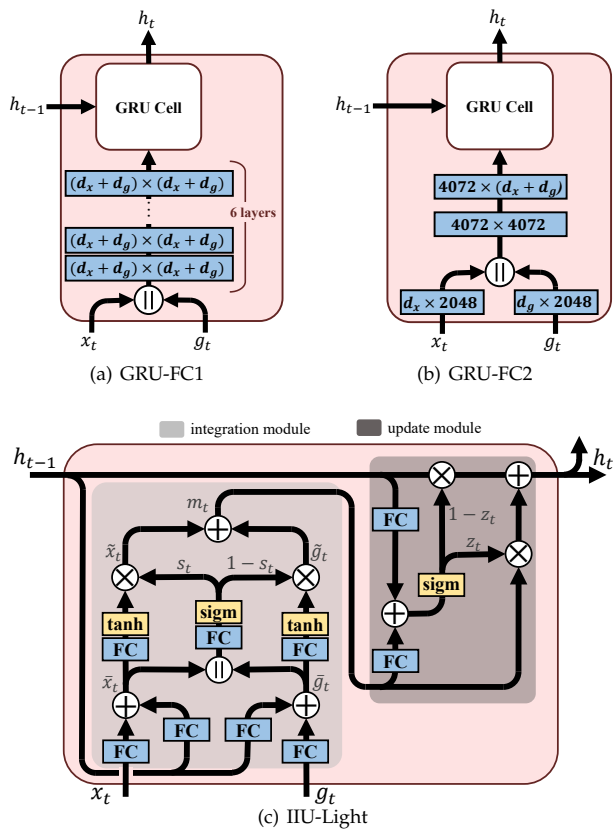

    \centering
    \hspace{-1.15em}
    \resizebox{\linewidth}{!}{
    \begin{tabular}{cc}
        \subfigure[GRU-FC1]{
            \includegraphics[width=0.46\linewidth]{figures/gru_fc_a.pdf}
        } &
        \subfigure[GRU-FC2]{
            \includegraphics[width=0.46\linewidth]{figures/gru_fc_b.pdf}
        } \\
        \multicolumn{2}{c}{\subfigure[IIU-Light]{
            {\includegraphics[width=1\linewidth]{figures/iiu-light.pdf}}
        }} \\
    \end{tabular}
    }
    \caption{
    Architectures of (a) GRU-FC1, (b) GRU-FC2, and (c) IIU-Light.
    }
    \label{fig:discussion}
\end{figure}
In Table~\ref{tab9}, we report the results on THUMOS-14~\cite{jiang2015url}.
Our IIN surpasses the state-of-the-art methods by a large margin at all future times.
IIN achieves $52.0\%$ mAP at 1 second into the future, 47.2\% mAP at 2 seconds into the future, and $51.2\%$ mAP on average.
Also, IIN-Kinetics outperforms all methods by a large margin by achieving $51.9\%$ mAP on average.
Through these results, we demonstrate the superiority of IIU in learning enriched features from the information on two different modalities.

We measure the inference time of IIN and TRN to compare the efficiencies.
TRN takes 92.6ms for action anticipation with the same size of hidden dimension and the same length of input videos as IIN, while our IIN takes 99.4ms as mentioned in Sec.~\ref{sec:iiu-ablation}.
For fair comparison, we modify TRN to take pseudo action labels as additional inputs. 
TRN with pseudo action labels takes 113.4ms, which is 14ms slower than IIN.
In this experiment, we also observe that using pseudo action labels improves the TRN performances by 0.2\% mcAP at 1 second and 1.3\% mcAP at 2 seconds on TVSeries, achieving 76.1\% mcAP and 73.6\% mcAP, respectively.


\subsubsection{Qualitative Evaluation}
We visualize the results of our IIN on TVSeries~\cite{geest2016eccv} and THUMOS-14~\cite{jiang2015url} datasets in Fig.~\ref{fig:aa} (a) and (b), respectively.
The results on TVSeries show that IIN correctly predicts the duration of the ongoing action and the next action classes.
In Fig.~\ref{fig:aa} (b), our IIN forecasts the starting time of a future action successfully, which means that IIN has the ability to recognize a scene change from an observed video as background.
Note that it is very challenging to anticipate the exact end time of unseen future actions.

\subsubsection{Discussion}
We conduct additional experiments to compare GRU baseline with the same amount of parameters to our IIU.
For the experiments, several Fully-Connected (FC) layers are added in front of GRU cell. 
We design two versions with different sizes of FC layers, named GRU-FC1 and GRU-FC2.
GRU-FC1 has 6 FC layers connected in series with a size of $(d_x + d_g) \times (d_x + d_g)$.
GRU-FC2 has one FC layers of size $d_x \times 2048$, $d_g \times 2048$, $4072 \times 4072$, and $4072 \times (d_x + d_g)$, respectively. 
Additionally, we introduce an IIU-Light, which is a variant of IIU with fewer parameters than IIU.
By modifying the update module of IIU, IIU-Light has 69.5\% of the number of parameters on IIU.
The architectures of GRU-FC1, GRU-FC2, and IIU-Light are illustrated in Fig.~\ref{fig:discussion}.



We summarize the results of GRU-FC1, GRU-FC2, and IIU-Light on TVSeries~\cite{geest2016eccv} in Table~\ref{tab:dis-tvseries}.
GRU-FC1 achieves 73.5\% mcAP at 1.0 seconds and 71.5\% mcAP at 2.0 seconds.
GRU-FC2 obtains 75.6\% mcAP and 71.5\% mcAP at 1.0 seconds and 2.0 seconds respectively, which are lower than the performances of our IIU.
IIU-Light achieves comparable performances to IIU with 76.9\% mcAP and 72.4\% mcAP at 1.0 and 2.0 seconds.
This result shows the superiority of the main idea of IIU, which is to exploit contextualized features from visual and action label features, on action anticipation.

In Table~\ref{tab:dis-thumos}, we report the results on THUMOS-14~\cite{jiang2015url}.
GRU-FC1 and GRU-FC2 achieve slight performance improvements compared to GRU, but do not outperform IIU.
Specifically, GRU-FC1 yields 48.8\% mAP and 40.2\% mAP at 1.0 and 2.0 seconds, respectively.
And, GRU-FC2 obtains 49.2\% mAP at 1.0 seconds and 40.7\% mAP at 2.0 seconds.
IIU-Light achieves 51.9\% mAP at 1.0 seconds and 46.7\% at 2.0 seconds, which are comparable performance to IIU.
\begin{table}[t!]
\centering
\caption{\label{tab:dis-tvseries} Performance comparisons for GRU-FC1, GRU-FC2, and IIU-Light on  TVSeries~\cite{geest2016eccv}. The best and the second best scores are marked in \textbf{bold}.}
\vspace{-0.3cm}
\centering
{\begin{tabular}{C{3cm} | C{2cm} C{2cm}}
\hline
\multirow{2}{*}{Method}           & \multicolumn{2}{c}{\begin{tabular}[c]{@{}c@{}}Time predicted into the future \\ (seconds)\end{tabular}} \\ \cline{2-3}
                   
                                  & 1.0s               & 2.0s               \\ \hline \hline
\multicolumn{1}{c|}{GRU}          & 73.9               & 71.2               \\ 
\multicolumn{1}{c|}{GRU-FC1}          &73.5                &71.5                \\
\multicolumn{1}{c|}{GRU-FC2}          &75.6            & 71.5               \\ 
\multicolumn{1}{c|}{IIU-Light}          &76.9             &72.4                \\ \cdashline{1-3}
\multicolumn{1}{c|}{IIU}          & \textbf{76.7}               & \textbf{74.3} \\ \hline
\end{tabular}
}
\end{table}
\begin{table}[t!]
\centering
\caption{\label{tab:dis-thumos} Performance comparisons for GRU-FC1, GRU-FC2, and IIU-Light on THUMOS-14~\cite{jiang2015url}. The best and the second best scores are marked in \textbf{bold}.}
\vspace{-0.3cm}
\centering
{\begin{tabular}{C{3cm} | C{2cm} C{2cm}}
\hline
\multirow{2}{*}{Method}           & \multicolumn{2}{c}{\begin{tabular}[c]{@{}c@{}}Time predicted into the future \\ (seconds)\end{tabular}} \\ \cline{2-3} 
                                  & 1.0s             & 2.0s               \\ \hline \hline
\multicolumn{1}{c|}{GRU}          & 48.4               & 40.0               \\ 
\multicolumn{1}{c|}{GRU-FC1}          &48.8                 &40.2               \\
\multicolumn{1}{c|}{GRU-FC2}          &49.2                &40.7                \\ 
\multicolumn{1}{c|}{IIU-Light}          &51.9                &46.7                \\ \cdashline{1-3}
\multicolumn{1}{c|}{IIU}          & \textbf{52.0}               & \textbf{47.2}  \\ \hline
\end{tabular}
}
\end{table}
\section{Conclusion}
In this paper, we proposed IDU that extends GRU \cite{cho2014emnlp} with two novel components: 1) a mechanism using current information and 2) an early embedding module.
These components enable IDU to effectively decide whether input information is relevant to a current action at every time step.
Based on IDU, our IDN effectively learns to discriminate relevant information from irrelevant information for identifying ongoing actions.
In comprehensive ablation studies, we demonstrated the generality and effectiveness of our proposed components.
Moreover, we confirmed that our IDN significantly outperforms state-of-the-art methods on TVSeries \cite{geest2016eccv} and THUMOS-14 \cite{jiang2015url} datasets for online action detection.

We further introduced IIU to demonstrate the applicability of our IDN for action anticipation.
Since pseudo action labels from IDN encourage the network to extract action-relevant information from visual features, IIU is able to exploit enriched features of observed actions.
According to the mechanism of IIU, our IIN captures sufficient contextual information for predicting future actions.
Experimental results show that our IIN achieves state-of-the-art performances on TVSeries and THUMOS-14 datasets for action anticipation.
Through these results, we empirically demonstrate the effectiveness of our proposed relation modeling with IDU and IIU. 
In the end, we show that our IDN has a great potential to be broadly applied beyond a single task (i.e., online action detection).


%



\section*{Acknowledgment}
This work was supported by Institute of Information \& Communications Technology Planning \& Evaluation (IITP) grant funded by the Korea government (MSIT) (No. 2020-0-00004, Development of Previsional Intelligence based on Long-term Visual Memory Network)




\bibliographystyle{IEEEtran}
\bibliography{mybibfile}

\begin{thebibliography}{10}
\providecommand{\url}[1]{#1}
\csname url@samestyle\endcsname
\providecommand{\newblock}{\relax}
\providecommand{\bibinfo}[2]{#2}
\providecommand{\BIBentrySTDinterwordspacing}{\spaceskip=0pt\relax}
\providecommand{\BIBentryALTinterwordstretchfactor}{4}
\providecommand{\BIBentryALTinterwordspacing}{\spaceskip=\fontdimen2\font plus
\BIBentryALTinterwordstretchfactor\fontdimen3\font minus
  \fontdimen4\font\relax}
\providecommand{\BIBforeignlanguage}[2]{{%
\expandafter\ifx\csname l@#1\endcsname\relax
\typeout{** WARNING: IEEEtran.bst: No hyphenation pattern has been}%
\typeout{** loaded for the language `#1'. Using the pattern for}%
\typeout{** the default language instead.}%
\else
\language=\csname l@#1\endcsname
\fi
#2}}
\providecommand{\BIBdecl}{\relax}
\BIBdecl

\bibitem{chao2018cvpr}
Y.-W. Chao, S.~Vijayanarasimhan, B.~Seybold, D.~A. Ross, J.~Deng, and
  R.~Sukthankar, ``Rethinking the faster r-cnn architecture for temporal action
  localization,'' in \emph{Proceedings of IEEE Conference on Computer Vision
  and Pattern Recognition}, Jun. 2018, pp. 1130--1139.

\bibitem{liu2019cvpr}
Y.~Liu, L.~Ma, Y.~Zhang, W.~Liu, and S.-F. Chang, ``Multi-granularity generator
  for temporal action proposal,'' in \emph{Proceedings of IEEE Conference on
  Computer Vision and Pattern Recognition}, Jun. 2019, pp. 3604--3613.

\bibitem{xu2017iccv}
H.~Xu, A.~Das, and K.~Saenko, ``R-c3d: Region convolutional 3d network for
  temporal activity detection,'' in \emph{Proceedings of IEEE International
  Conference on Computer Vision (ICCV)}, Oct. 2017, pp. 5783--5792.

\bibitem{zhang2019aaai}
X.-Y. Zhang, H.~Shi, C.~Li, X.~Z. K.~Zheng, and L.~Duan, ``Learning
  transferable self-attentive representations for action recognition in
  untrimmed videos with weak supervision,'' in \emph{Proceedings of Association
  for the Advancement of Artificial Intelligence (AAAI) Conference on
  Artificial Intelligence}, Jan. 2019, pp. 9227--9242.

\bibitem{zhao2017iccv}
Y.~Zhao, Y.~Xiong, L.~Wang, Z.~Wu, X.~Tang, and D.~Lin, ``Temporal action
  detection with structured segment networks,'' in \emph{Proceedings of IEEE
  Conference on Computer Vision and Pattern Recognition}, Oct. 2017, pp.
  2914--2923.

\bibitem{kim2019cvpr}
J.~Kim, T.~Misu, Y.-T. Chen, A.~Tawari, and J.~Canny, ``Grounding
  human-to-vehicle advice for self-driving vehicles,'' in \emph{Proceedings of
  IEEE Conference on Computer Vision and Pattern Recognition}, Jun. 2019, pp.
  10\,591--10\,599.

\bibitem{koppula2013iros}
H.~S. Koppula and A.~Saxena, ``Anticipating human activities for reactive
  robotic response,'' in \emph{Proceedings of International Conference on
  Intelligent Robots and Systems (IROS)}, Nov. 2013, pp. 2071--2071.

\bibitem{iwashita2013bmvc}
Y.~Iwashita, M.~Ryoo, T.~J. Fuchs, and C.~Padgett, ``Recognizing humans in
  motion: Trajectory-based aerial video analysis,'' in \emph{Proceedings of
  British Machine Vision Conference}, Sep. 2013, pp. 127.1--127.11.

\bibitem{shu2015cvpr}
T.~Shu, D.~Xie, B.~Rothrock, S.~Todorovic, and S.~C. Zhu, ``Joint inference of
  groups, events and human roles in aerial videos,'' in \emph{Proceedings of
  IEEE Conference on Computer Vision and Pattern Recognition}, Jun. 2015, pp.
  4756--4584.

\bibitem{cho2014emnlp}
K.~Cho, B.~van Merri{\"e}nboer, C.~Gulcehre, D.~Bahdanau, F.~Bougares,
  H.~Schwenk, and Y.~Bengio, ``Learning phrase representations using rnn
  encoder-decoder for statistical machine translation,'' in \emph{Proceedings
  of Conference on Empirical Methods in Natural Language Processing (EMNLP)},
  Oct. 2014, pp. 1724--1734.

\bibitem{gao22017bmvc}
J.~Gao, Z.~Yang, and R.~Nevatia, ``Red: Reinforced encoder-decoder networks for
  action anticipation,'' in \emph{Proceedings of British Machine Vision
  Conference}, Sep. 2017, pp. 92.1--92.11.

\bibitem{xu2019iccv}
M.~Xu, M.~Gao, Y.-T. Chen, L.~S. Davis, and D.~J. Crandall, ``Temporal
  recurrent networks for online action detection,'' in \emph{Proceedings of
  IEEE International Conference on Computer Vision (ICCV)}, Oct. 2019, pp.
  5532--5541.

\bibitem{hochreiter1997nc}
S.~Hochreiter and J.~Schmidhuber, ``Long short-term memory,'' \emph{Neural
  Computation}, vol.~9, pp. 1735--1780, Dec. 1997.

\bibitem{geest2016eccv}
R.~D. Geest, E.~Gavves, A.~Ghodrati, Z.~Li, G.~Snoek, and T.~Tuytelaars,
  ``Online action detection,'' in \emph{Proceedings of European Conference on
  Computer Vision}, Oct. 2016, pp. 269--285.

\bibitem{jiang2015url}
Y.~G. Jiang, J.~Liu, A.~R. Zamir, G.~Toderici, I.~Laptev, M.~Shah, and
  R.~Sukthankar, ``Thumos challenge: Action recognition with a large number of
  classes,'' 2014, http://crcv.ucf.edu/THUMOS14/.

\bibitem{eun2020learning}
H.~Eun, J.~Moon, J.~Park, C.~Jung, and C.~Kim, ``Learning to discriminate
  information for online action detection,'' in \emph{Proceedings of IEEE
  Conference on Computer Vision and Pattern Recognition}, 2020, pp. 809--818.

\bibitem{shou2016cvpr}
Z.~Shou, D.~Wang, and S.-F. Chang, ``Temporal action localization in untrimmed
  videos via multi-stage cnns,'' in \emph{Proceedings of IEEE Conference on
  Computer Vision and Pattern Recognition}, Jun. 2016, pp. 1049--1058.

\bibitem{dai2017iccvr}
X.~Dai, B.~Singh, G.~Zhang, L.~S. Davis, and Y.~Q. Chen, ``Temporal context
  network for activity localization in videos,'' in \emph{Proceedings of IEEE
  International Conference on Computer Vision (ICCV)}, Oct. 2017, pp.
  5727--5736.

\bibitem{shou2017cvpr}
Z.~Shou, J.~Chan, A.~Zareian, K.~Miyazawa, and S.-F. Chang, ``Cdc:
  Convolutional-de-convolutional networks for precise temporal action
  localization in untrimmed videos,'' in \emph{Proceedings of IEEE Conference
  on Computer Vision and Pattern Recognition}, Jul. 2017, pp. 5734--5743.

\bibitem{r2-3}
R.~Su, D.~Xu, L.~Sheng, and W.~Ouyang, ``Pcg-tal: Progressive cross-granularity
  cooperation for temporal action localization,'' \emph{IEEE Transactions on
  Image Process.}, vol.~30, pp. 2103--2113, 2020.

\bibitem{r2-2}
C.~Lin, C.~Xu, D.~Luo, Y.~Wang, Y.~Tai, C.~Wang, J.~Li, F.~Huang, and Y.~Fu,
  ``Learning salient boundary feature for anchor-free temporal action
  localization,'' in \emph{Proceedings of IEEE Conference on Computer Vision
  and Pattern Recognition}, 2021, pp. 3320--3329.

\bibitem{r2-6}
R.~Su, D.~Xu, L.~Zhou, and W.~Ouyang, ``Improving weakly supervised temporal
  action localization by exploiting multi-resolution information in temporal
  domain,'' \emph{IEEE Transactions on Image Process.}, vol.~30, pp.
  6659--6672, 2021.

\bibitem{r2-1}
------, ``Progressive cross-stream cooperation in spatial and temporal domain
  for action localization,'' \emph{IEEE Transactions on Pattern Analysis and
  Machine Intelligence}, vol.~43, no.~12, pp. 4477--4490, 2020.

\bibitem{r2-4}
L.~Wang, Z.~Tong, B.~Ji, and G.~Wu, ``Tdn: Temporal difference networks for
  efficient action recognition,'' in \emph{Proceedings of IEEE Conference on
  Computer Vision and Pattern Recognition}, 2021, pp. 1895--1904.

\bibitem{r2-7}
X.~Wang, S.~Zhang, Z.~Qing, Y.~Shao, C.~Gao, and N.~Sang, ``Self-supervised
  learning for semi-supervised temporal action proposal,'' in \emph{Proceedings
  of IEEE Conference on Computer Vision and Pattern Recognition}, 2021, pp.
  1905--1914.

\bibitem{r2-5}
Z.~Qing, H.~Su, W.~Gan, D.~Wang, W.~Wu, X.~Wang, Y.~Qiao, J.~Yan, C.~Gao, and
  N.~Sang, ``Temporal context aggregation network for temporal action proposal
  refinement,'' in \emph{Proceedings of IEEE Conference on Computer Vision and
  Pattern Recognition}, 2021, pp. 485--494.

\bibitem{ren2015nips}
S.~Ren, K.~He, R.~Girchick, and J.~Sun, ``Faster r-cnn: Towards real-time
  object detection with region proposal networks,'' in \emph{Proceedings of
  Advances in Neural Information Processing Systems}, Dec. 2015, pp. 91--99.

\bibitem{3r_iccv}
L.~Huang, L.~Wang, and H.~Li, ``Foreground-action consistency network for
  weakly supervised temporal action localization,'' in \emph{Proceedings of
  IEEE International Conference on Computer Vision (ICCV)}, 2021, pp.
  8002--8011.

\bibitem{3r_aaai}
L.~Huang, Y.~Huang, W.~Ouyang, and L.~Wang, ``Relational prototypical network
  for weakly supervised temporal action localization,'' in \emph{Proceedings of
  Association for the Advancement of Artificial Intelligence (AAAI) Conference
  on Artificial Intelligence}, vol.~34, no.~07, 2020, pp. 11\,053--11\,060.

\bibitem{donahue2015cvpr}
J.~Donahue, L.~A. Hendricks, S.~Guadarrama, M.~Rohrbach, S.~Venugopalan,
  K.~Saenko, and T.~Darrell, ``Long-term recurrent convolutional networks for
  visual recognition and description,'' in \emph{Proceedings of IEEE Conference
  on Computer Vision and Pattern Recognition}, Jun. 2015, pp. 2625--2634.

\bibitem{yeung2018ijcv}
S.~Yeung, O.~Russakovsky, N.~Jin, M.~Andriluka, G.~Mori, and L.~Fei-Fei,
  ``Every moment counts: Dense detailed labeling of actions in complex
  videos,'' \emph{International Journal of Computer Vision}, vol. 126, pp.
  375--389, Apr. 2018.

\bibitem{hoai2012cvpr}
M.~Hoai and F.~D. la~Torre, ``Max-margin early event detectors,'' in
  \emph{Proceedings of IEEE Conference on Computer Vision and Pattern
  Recognition}, Jun. 2012, pp. 2863--2870.

\bibitem{hoai2014ijcv}
------, ``Max-margin early event detector,'' \emph{International Journal of
  Computer Vision}, vol.~2, pp. 191--202, Apr. 2014.

\bibitem{ma2016cvpr}
S.~Ma, L.~Signal, and S.~Sclaroff, ``Learning activity progression in lstms for
  activity detection and early detection,'' in \emph{Proceedings of IEEE
  Conference on Computer Vision and Pattern Recognition}, Jun. 2016, pp.
  1942--1950.

\bibitem{early1}
H.~Gammulle, S.~Denman, S.~Sridharan, and C.~Fookes, ``Predicting the future: A
  jointly learnt model for action anticipation,'' in \emph{Proceedings of IEEE
  International Conference on Computer Vision (ICCV)}, 2019, pp. 5562--5571.

\bibitem{early2}
J.-F. Hu, W.-S. Zheng, L.~Ma, G.~Wang, J.~Lai, and J.~Zhang, ``Early action
  prediction by soft regression,'' \emph{IEEE Transactions on Pattern Analysis
  and Machine Intelligence}, vol.~41, no.~11, pp. 2568--2583, 2018.

\bibitem{early3}
Y.~Kong, Z.~Tao, and Y.~Fu, ``Deep sequential context networks for action
  prediction,'' in \emph{Proceedings of IEEE Conference on Computer Vision and
  Pattern Recognition}, 2017, pp. 1473--1481.

\bibitem{early4}
M.~Sadegh~Aliakbarian, F.~Sadat~Saleh, M.~Salzmann, B.~Fernando, L.~Petersson,
  and L.~Andersson, ``Encouraging lstms to anticipate actions very early,'' in
  \emph{Proceedings of IEEE International Conference on Computer Vision
  (ICCV)}, 2017, pp. 280--289.

\bibitem{tsochantaridis2005jmlr}
I.~Tsochantaridis, T.~Joachims, T.~Hofmann, and Y.~Altun, ``Large margin
  methods for structured and interdependent output variables,'' \emph{Journal
  of Machine Learning Research (JMLR)}, vol.~6, pp. 1453--1484, Sep. 2005.

\bibitem{cai2019aaai}
Y.~Cai, H.~Li, J.-F. Hu, and W.-S. Zheng, ``Action knowledge transfer for
  action prediction with partial videos,'' in \emph{Proceedings of Association
  for the Advancement of Artificial Intelligence (AAAI) Conference on
  Artificial Intelligence}, Jan. 2019, pp. 8118--8125.

\bibitem{geest2018wacv}
R.~D. Geest and T.~Tuytelaars, ``Modeling temporal structure with lstm for
  online action detection,'' in \emph{Proceedings of IEEE Winter Conference on
  Applications of Computer Vision}, Mar. 2018, pp. 1549--1557.

\bibitem{3r_nips}
M.~Xu, Y.~Xiong, H.~Chen, X.~Li, W.~Xia, Z.~Tu, and S.~Soatto, ``Long
  short-term transformer for online action detection,'' \emph{Advances in
  Neural Information Processing Systems}, vol.~34, pp. 1086--1099, 2021.

\bibitem{next1}
T.~Lan, T.-C. Chen, and S.~Savarese, ``A hierarchical representation for future
  action prediction,'' in \emph{Proceedings of European Conference on Computer
  Vision}.\hskip 1em plus 0.5em minus 0.4em\relax Springer, 2014, pp. 689--704.

\bibitem{next2}
T.~Mahmud, M.~Hasan, and A.~K. Roy-Chowdhury, ``Joint prediction of activity
  labels and starting times in untrimmed videos,'' in \emph{Proceedings of IEEE
  International Conference on Computer Vision (ICCV)}, 2017, pp. 5773--5782.

\bibitem{next4}
N.~Rhinehart and K.~M. Kitani, ``First-person activity forecasting with online
  inverse reinforcement learning,'' in \emph{Proceedings of IEEE International
  Conference on Computer Vision (ICCV)}, 2017, pp. 3696--3705.

\bibitem{next3}
H.~S. Koppula and A.~Saxena, ``Anticipating human activities using object
  affordances for reactive robotic response,'' \emph{IEEE Transactions on
  Pattern Analysis and Machine Intelligence}, vol.~38, no.~1, pp. 14--29, 2015.

\bibitem{next5}
M.~Pei, Z.~Si, B.~Z. Yao, and S.-C. Zhu, ``Learning and parsing video events
  with goal and intent prediction,'' \emph{Computer Vision and Image
  Understanding}, vol. 117, no.~10, pp. 1369--1383, 2013.

\bibitem{next6}
A.~Jain, A.~R. Zamir, S.~Savarese, and A.~Saxena, ``Structural-rnn: Deep
  learning on spatio-temporal graphs,'' in \emph{Proceedings of IEEE Conference
  on Computer Vision and Pattern Recognition}, 2016, pp. 5308--5317.

\bibitem{3r_tip}
H.~Wang, J.~Dong, B.~Cheng, and J.~Feng, ``Pvred: a position-velocity recurrent
  encoder-decoder for human motion prediction,'' \emph{IEEE Transactions on
  Image Process.}, vol.~30, pp. 6096--6106, 2021.

\bibitem{fewsec1}
C.~Vondrick, H.~Pirsiavash, and A.~Torralba, ``Anticipating visual
  representations from unlabeled video,'' in \emph{Proceedings of IEEE
  Conference on Computer Vision and Pattern Recognition}, 2016, pp. 98--106.

\bibitem{fewsec5}
S.~Qi, S.~Huang, P.~Wei, and S.-C. Zhu, ``Predicting human activities using
  stochastic grammar,'' in \emph{Proceedings of IEEE International Conference
  on Computer Vision (ICCV)}, 2017, pp. 1164--1172.

\bibitem{ego2}
E.~Dessalene, C.~Devaraj, M.~Maynord, C.~Fermuller, and Y.~Aloimonos,
  ``Forecasting action through contact representations from first person
  video,'' \emph{IEEE Transactions on Pattern Analysis and Machine
  Intelligence}, 2021.

\bibitem{ego5}
E.~Dessalene, M.~Maynord, C.~Devaraj, C.~Fermuller, and Y.~Aloimonos,
  ``Egocentric object manipulation graphs,'' \emph{arXiv preprint
  arXiv:2006.03201}, 2020.

\bibitem{fewsec4}
A.~Furnari and G.~M. Farinella, ``What would you expect? anticipating
  egocentric actions with rolling-unrolling lstms and modality attention,'' in
  \emph{Proceedings of IEEE International Conference on Computer Vision
  (ICCV)}, 2019, pp. 6252--6261.

\bibitem{ego3}
N.~Osman, G.~Camporese, P.~Coscia, and L.~Ballan, ``Slowfast rolling-unrolling
  lstms for action anticipation in egocentric videos,'' in \emph{Proceedings of
  IEEE International Conference on Computer Vision (ICCV)}, 2021, pp.
  3437--3445.

\bibitem{ego4}
B.~Fernando and S.~Herath, ``Anticipating human actions by correlating past
  with the future with jaccard similarity measures,'' in \emph{Proceedings of
  IEEE Conference on Computer Vision and Pattern Recognition}, 2021, pp.
  13\,224--13\,233.

\bibitem{ego1}
Q.~Ke, M.~Fritz, and B.~Schiele, ``Time-conditioned action anticipation in one
  shot,'' in \emph{Proceedings of IEEE Conference on Computer Vision and
  Pattern Recognition}, 2019, pp. 9925--9934.

\bibitem{ego6}
Y.~Abu~Farha, A.~Richard, and J.~Gall, ``When will you do what?-anticipating
  temporal occurrences of activities,'' in \emph{Proceedings of IEEE Conference
  on Computer Vision and Pattern Recognition}, 2018, pp. 5343--5352.

\bibitem{nari2010icml}
V.~Nair and G.~E. Hinton, ``Rectified linear units improve restricted obltzmann
  machines,'' in \emph{Proceedings of International Conference on Machine
  Learning}, Jun. 2010.

\bibitem{chopra2005cvpr}
S.~Chopra, R.~Hadsell, and Y.~LeCun, ``Learning a similarity metric
  discriminatively, with application to face verification,'' in
  \emph{Proceedings of IEEE Conference on Computer Vision and Pattern
  Recognition}, Jun. 2005, pp. 539--546.

\bibitem{hadsell2006cvpr}
R.~Hadsell, S.~Chopra, and Y.~LeCun, ``Dimensionality reduction by learning an
  invariant mapping,'' in \emph{Proceedings of IEEE Conference on Computer
  Vision and Pattern Recognition}, Jun. 2006, pp. 1735--1742.

\bibitem{sohn2016nips}
K.~Sohn, ``Improved deep metric learning with multi-class n-pair loss
  objective,'' in \emph{Proceedings of Advances in Neural Information
  Processing Systems}, Dec. 2016, pp. 1857--1865.

\bibitem{wang2016eccv}
L.~Wang, Y.~Xiong, Z.~Wang, Y.~Q.~D. Lin, X.~Tang, and L.~van Gool, ``Temporal
  segment networks: Towards good practices for deep action recognition,'' in
  \emph{Proceedings of European Conference on Computer Vision}, Oct. 2016, pp.
  20--36.

\bibitem{heilbron2015cvpr}
F.~C. Heilbron, B.~G. V.~Escorcia, and J.~C. Niebles, ``Activitynet: A
  large-scale video benchmark for human activity understanding,'' in
  \emph{Proceedings of IEEE Conference on Computer Vision and Pattern
  Recognition}, Jun. 2015, pp. 961--970.

\bibitem{he2016cvpr}
K.~He, X.~Zhang, S.~Ren, and J.~Sun, ``Deep residual learning for image
  recognition,'' in \emph{Proceedings of IEEE Conference on Computer Vision and
  Pattern Recognition}, Jul. 2016, pp. 771--778.

\bibitem{ioffe2015arxiv}
S.~Ioffe and C.~Szegedy, ``Batch normalization: Accelerating deep network
  training by reducing internal covariate shift,'' \emph{arXiv:1502.03167},
  2015.

\bibitem{adam}
D.~P. Kingma and J.~Ba, ``Adam: {A} method for stochastic optimization,'' in
  \emph{Proceedings of International Conference on Learning Representations},
  2015.

\bibitem{carreira2017cvpr}
J.~Carreira and A.~Zisserman, ``Quo vaids, action recognition? a new model and
  the kinectics dataset,'' in \emph{Proceedings of IEEE Conference on Computer
  Vision and Pattern Recognition}, Jul. 2017, pp. 4724--4733.

\bibitem{simonyan2015iclr}
K.~Simonyan and A.~Zisserman, ``Very deep convolutional networks for
  large-scale image recognition,'' in \emph{Proceedings of International
  Conference on Learning Representations}, May 2015.

\bibitem{simonyan2014nips}
------, ``Two-stream convolutional networks for action recognition in videos,''
  in \emph{Proceedings of Advances in Neural Information Processing Systems},
  Dec. 2014, pp. 568--576.

\end{thebibliography}
%



%

\begin{IEEEbiography}[{\includegraphics[width=1in,height=1.25in,clip,keepaspectratio]{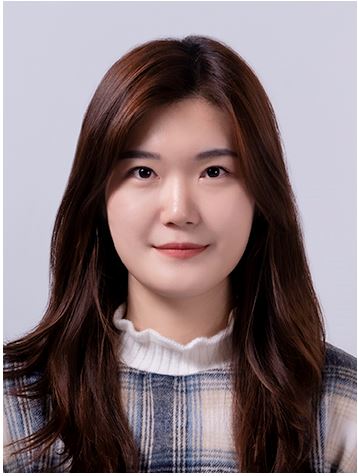}}]{Sumin Lee}
received the B.S. degree in the School of Electronic engineering from Kyungpook National University, Daegu, South Korea, in 2018, and the M.S. degree in the school of electrical engineering from Korea Advanced Institue of Science and Technology  (KAIST), Daejeon, South Korea, in 2020. She is currently pursuing the Ph.D. degree in electrical engineering with the school of electrical engineering from KAIST.
Her research interest includes action detection, anticipation, and localization for video understanding.
\end{IEEEbiography}
\begin{IEEEbiography}[{\includegraphics[width=1in,height=1.25in,clip,keepaspectratio]{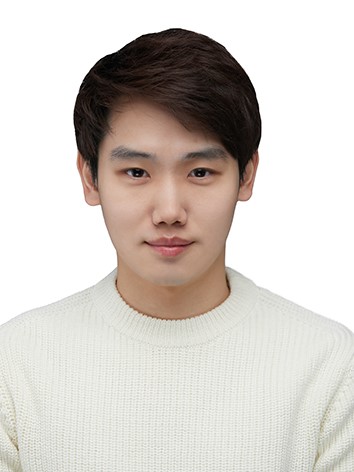}}]{Hyunjun Eun}
received the B.S. degree in electronic engineering from the Kyungpook National University (KNU), Daegu, South Korea, in 2013, and the M.S. and Ph.D. degrees in electrical engineering the from the Korea Advanced Institute of Science and Technology (KAIST), Daejeon, South Korea, in 2015 and 2020, respectively. Since 2020, he has been working with the Video Recognition Tech. Cell, AI Service Division, SK Telecom, Seoul, South Korea. His current research interests include action detection and recognition for video understanding and text detection.
\end{IEEEbiography}
\begin{IEEEbiography}[{\includegraphics[width=1in,height=1.25in,clip,keepaspectratio]{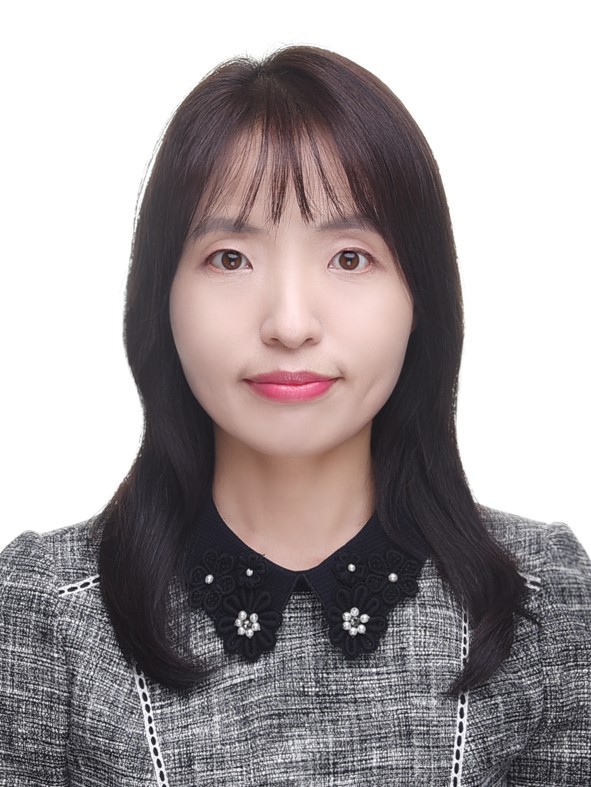}}]{Jinyoung Moon}
received her B.S. degree in Computer Engineering from the Kyungpook National University (KNU), Daegu, Rep. of Korea, in 2000. She received her M.S. degree in Computer Science and Ph.D. in Industrial \& Systems Engineering from the Korea Advanced Institute of Science and Technology (KAIST), Daejeon, Rep. of Korea, in 2002 and 2018, respectively. Since 2002, she has been working with the Visual Intelligence Research Section, the Artificial Intelligence Research Laboratory, the Electronics and Telecommunications Research Institute (ETRI), Daejeon, Rep. of Korea. Since 2019, she has also been with the ICT department, the University of Science and Technology (UST), where she is currently an Assistant Professor. Her research interests include action recognition, action detection, temporal moment localization, and video QA.
\end{IEEEbiography}
\begin{IEEEbiography}[{\includegraphics[width=1in,height=1.25in,clip,keepaspectratio]{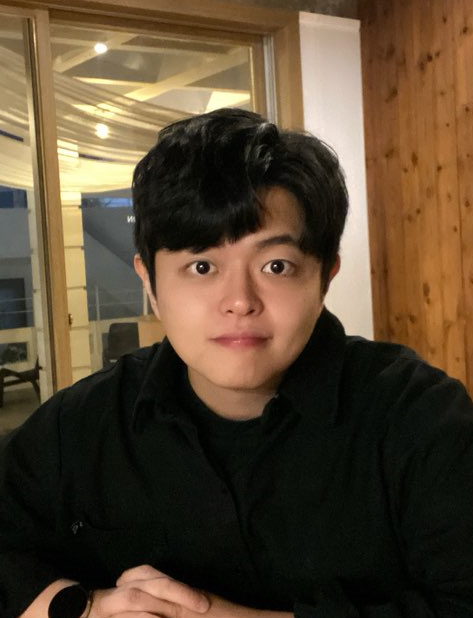}}]{Seokeon Choi}
received the B.S. degree in the school of electronic and electrical engineering from Sungkyunkwan University, Suwon, South Korea, in 2015, and the M.S. degree in the school of electrical engineering from Korea Advanced Institute of Science and Technology (KAIST), Daejeon, South Korea, in 2017. Currently, he is a Ph.D. candidate in the school of electrical engineering from KAIST. From January 2020 to July 2020, he was a Visiting Student in the Department of Language Technologies Institute from Carnegie Mellon University. His research interests are computer vision and machine learning, with an emphasis on person re-identification, object tracking, domain generalization, human-oriented visual understanding, and machine perception. 
\end{IEEEbiography}
\begin{IEEEbiography}[{\includegraphics[width=1in,height=1.25in,clip,keepaspectratio]{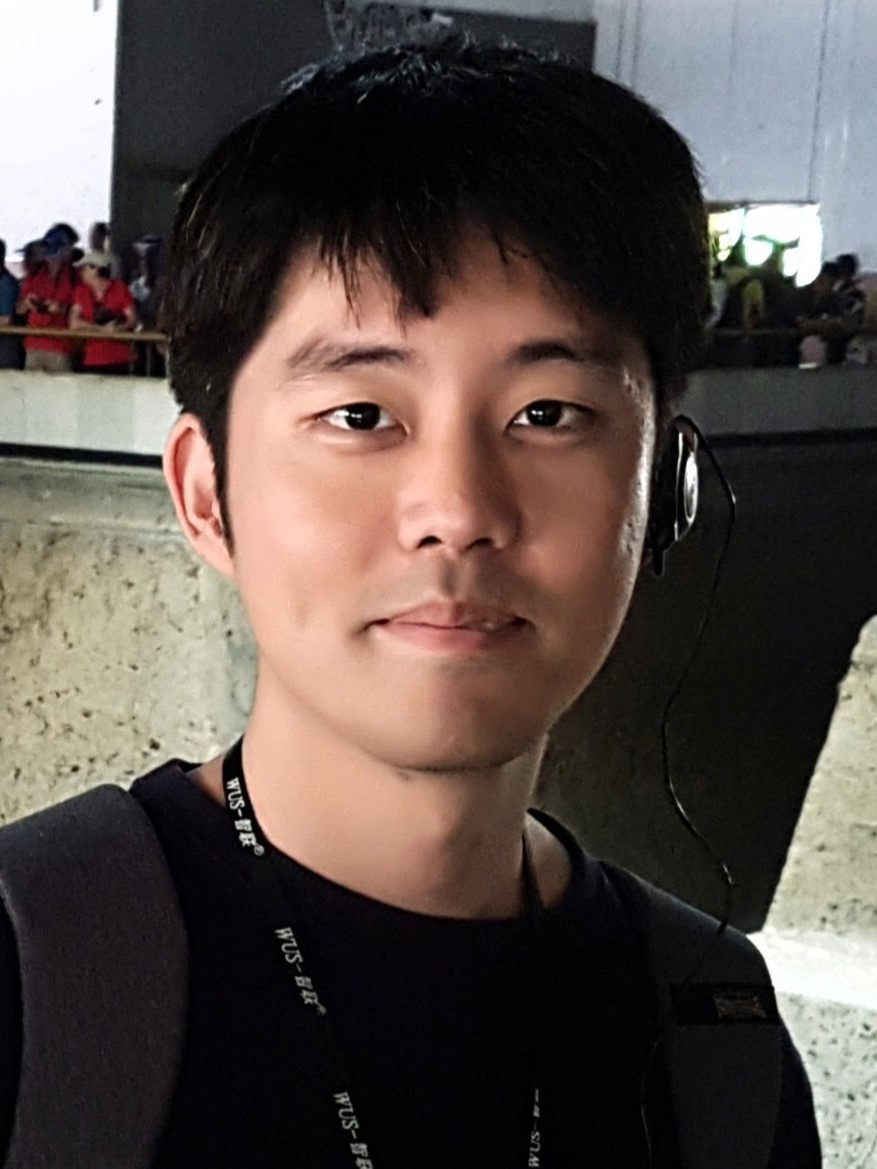}}]{Yoonhyung Kim} 
received the B.S., M.S., and Ph. D. degrees in electrical engineering from Korea Advanced Institute of Science and Technology (KAIST), Daejeon, Republic of Korea, in 2013, 2016, and 2021, respectively. Since 2021, he has been working with the Artificial Intelligence Research Laboratory, Electronics and Telecommunications Research Institute (ETRI), Daejeon, Republic of Korea. His current research interests include computer vision, speech recognition, and multi-modal deep learning.
\end{IEEEbiography}
\begin{IEEEbiography}[{\includegraphics[width=1in,height=1.25in,clip,keepaspectratio]{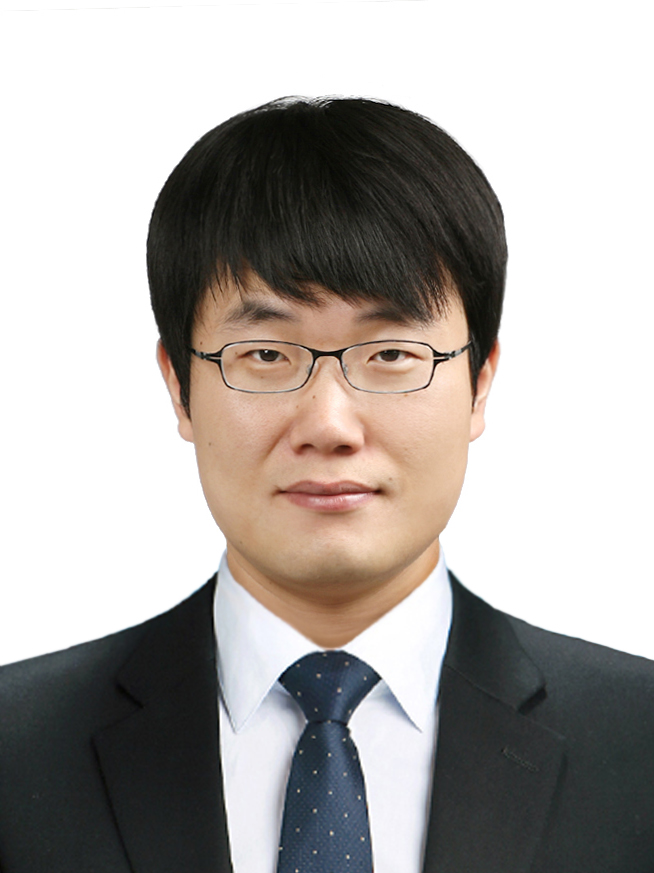}}]{Chanho Jung}
received the B.S. and M.S. degrees in electronic engineering from Sogang University, Seoul, South Korea, in 2004 and 2006, respecti vely, and the Ph.D. degree in electrical engineering from the Korea Advanced Institute of Science and Technology (KAIST), Daejeon, South Korea, in 2013. From 2006 to 2008, he was a Research Engineer with the Digital Television Research Laboratory, LG Electronics, Seoul. From 2013 to 2016, he was a Senior Researcher with the Electronics and Telecommunications Research Institute (ETRI), Daejeon. Since 2016, he has been with the Department of Electrical Engineering, Hanbat National University, Daejeon, where he is currently an Associate Professor. His current research interests include computer vision, machine learning, embedded systems, pattern recognition, and image processing.
\end{IEEEbiography}
\begin{IEEEbiography}[{\includegraphics[width=1in,height=1.25in,clip,keepaspectratio]{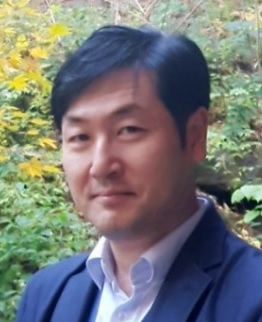}}]{Changick Kim}
received the B.S. degree in electrical engineering from Yonsei University, Seoul, South Korea, in 1989, the M.S. degree in electronics and electrical engineering from the Pohang University of Science and Technology (POSTECH), Pohang, South Korea, in 1991, and the Ph.D. degree in electrical engineering from the University of Washington, Seattle, WA, USA, in 2000. From 2000 to 2005, he was a Senior Member of Technical Staff with Epson Research and Development, Inc., Palo Alto, CA, USA. From 2005 to 2009, he was an Associate Professor with the School of Engineering, Information and Communications University, Daejeon, South Korea. Since March 2009, he has been with the School of Electrical Engineering, Korea Advanced Institute of Science and Technology (KAIST), Daejeon, Korea, where he is currently a Professor. He is also in charge of the center for security technology research, KAIST. His research interests include few shop learning, adversarial attack, and 3D reconstruction
\end{IEEEbiography}







\end{document}